\documentclass[nohyperref]{article}

\usepackage{microtype}
\usepackage{graphicx}
\usepackage{subfigure}
\usepackage{booktabs} 

\PassOptionsToPackage{hyphens}{url}\usepackage{hyperref}

\usepackage[accepted]{icml2022}

\usepackage{amsmath}
\usepackage{amssymb}
\usepackage{mathtools}
\usepackage{amsthm}

\usepackage[capitalize,noabbrev]{cleveref}

\theoremstyle{plain}

\theoremstyle{definition}

\theoremstyle{remark}

\newcommand{\x}{\mathbf{x}}

\newcommand{\bx}{\mathbf{x}}

\usepackage[textsize=tiny]{todonotes}

\usepackage{url}            
\usepackage{booktabs}       
\usepackage{amsfonts}       
\usepackage{nicefrac}       
\usepackage{microtype}      
\usepackage{xcolor}         

\usepackage{graphicx}
\usepackage{subfigure}
\usepackage{amsmath}

\usepackage{wrapfig}
\usepackage[toc,page]{appendix}

\usepackage{amsmath,amsfonts,bm}

\def\eqref#1{equation~\ref{#1}}

\def\1{\bm{1}}

\DeclareMathAlphabet{\mathsfit}{\encodingdefault}{\sfdefault}{m}{sl}
\SetMathAlphabet{\mathsfit}{bold}{\encodingdefault}{\sfdefault}{bx}{n}

\def\gL{{\mathcal{L}}}

\def\bx{\mathbf{x}}

\newcommand{\E}{\mathbb{E}}

\newcommand{\design}{\mathbf{x}}
\newcommand{\score}{y}
\newcommand{\dataset}{\mathcal{D}}
\newcommand{\oracle}{f}
\newcommand{\optima}{\design^*}
\newcommand{\algo}{\mathfrak{A}}

\icmltitlerunning{Design-Bench: Benchmarks for Data-Driven Offline Model-Based Optimization}

\begin{document}

\twocolumn[
\icmltitle{Design-Bench: Benchmarks for Data-Driven Offline Model-Based Optimization}

\icmlsetsymbol{equal}{*}

\begin{center}
\textbf{Brandon Trabucco}$^{*, \phi}$~~ \textbf{Xinyang Geng}$^{*, \theta}$~~ \textbf{Aviral Kumar}$^{\theta}$~~ \textbf{Sergey Levine}$^\theta$\\
$^\phi$CMU~~~~ $^\theta$UC Berkeley ~~~~~~ ($^*$Equal Contribution)\\
\texttt{btrabucco@berkeley.edu, young.geng@berkeley.edu}
\end{center}

\vskip 0.3in

\begin{abstract}
Black-box model-based optimization (MBO) problems,
where the goal is to find a design input that maximizes an unknown objective function,
are ubiquitous in a wide range of domains, such as the design of proteins, DNA sequences, aircraft, and robots.
Solving model-based optimization problems typically requires actively querying the unknown objective function on design proposals, which means physically building the candidate molecule, aircraft, or robot, testing it, and storing the result. This process can be expensive and time consuming, and one might instead prefer to optimize for the best design using only the data one already has. This setting---called offline MBO---poses substantial and different algorithmic challenges than more commonly studied online techniques.
A number of recent works have demonstrated success with offline MBO for high-dimensional optimization problems using high-capacity deep neural networks.
However, the lack of standardized benchmarks in this emerging field is making progress difficult to track. To address this, we present Design-Bench, a benchmark for offline MBO with a unified evaluation protocol and reference implementations of recent methods. Our benchmark includes a suite of diverse and realistic tasks derived from real-world optimization problems in biology, materials science, and robotics that present distinct challenges for offline MBO. Our benchmark and reference implementations are released at \href{https://github.com/rail-berkeley/design-bench}{github.com/rail-berkeley/design-bench} and \href{https://github.com/rail-berkeley/design-baselines}{github.com/rail-berkeley/design-baselines}.
\end{abstract}
]

\section{Introduction}
Automatically synthesizing designs that maximize a desired objective function is one of the most important challenges in scientific and engineering disciplines. From protein design in molecular biology~\cite{Shen2014} to superconducting material discovery in physics~\cite{hamidieh2018superconductor}, researchers have made significant progress in applying machine learning to optimization problems over structured design spaces.

Commonly, the exact form of the objective function is unknown, and the objective value for a novel design can only be found by either running computer simulations or real world experiments. This process of optimizing an unknown function by only observing samples from this function is known as black-box optimization, and is typically solved in an \textbf{online} iterative manner, where in each iteration the solver proposes new designs and queries the objective function for feedback in order to inform better design proposals at the next iteration~\cite{williams2006gaussian}. In many domains however, the objective function is prohibitively expensive to evaluate because it requires manually conducting experiments in the real world. In this setting, one cannot query the true objective function, and cannot receive feedback on design proposals. Instead, a collection of past records of designs and corresponding objective values might be available, and the optimization method must instead leverage existing data to synthesize the most optimal designs. This is called \textbf{offline model-based optimization} (offline MBO).

\begin{figure}[h]
    \centering
    \vspace{-0.2cm}
    \includegraphics[width=0.99\linewidth]{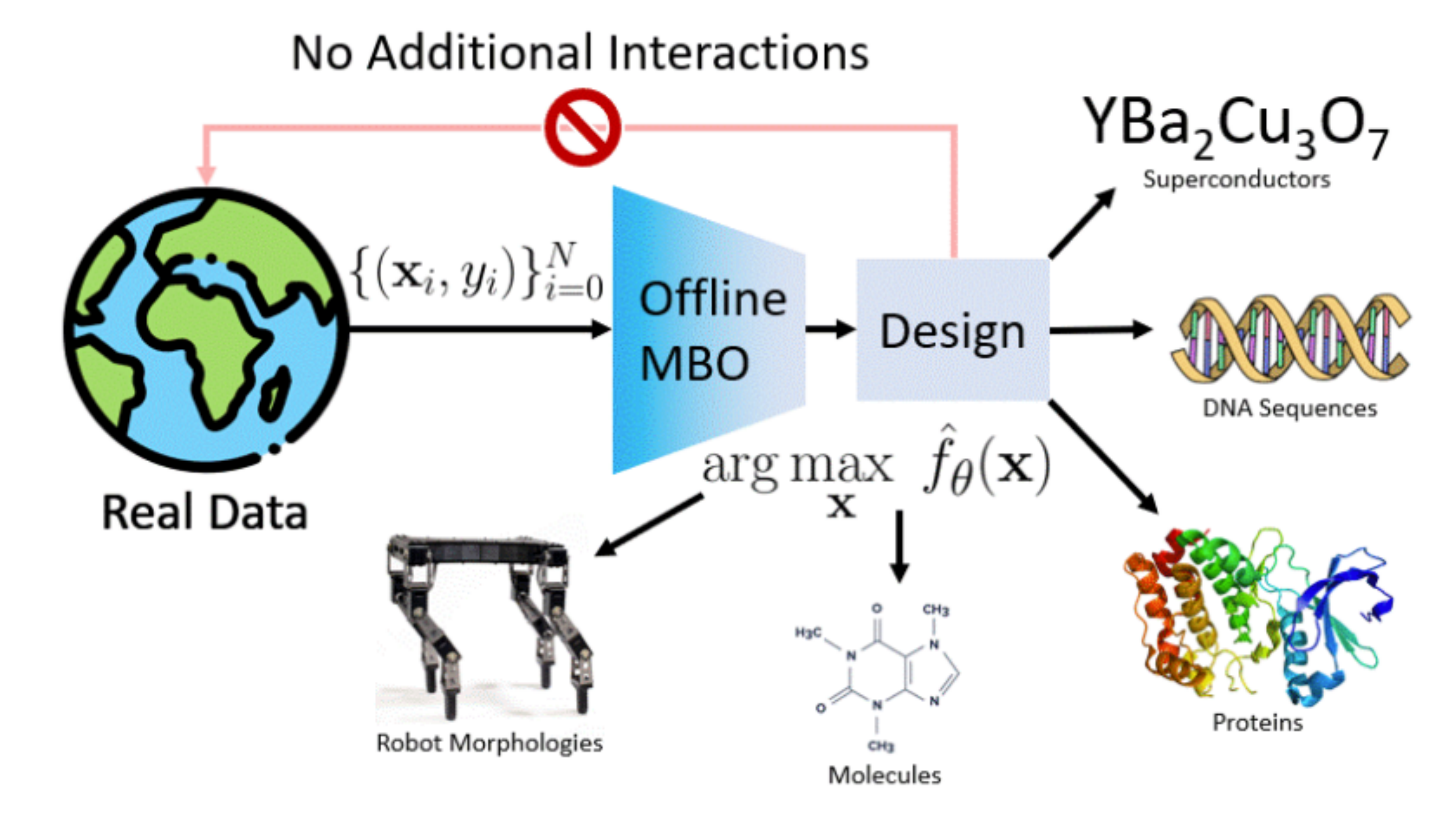}
    \vspace{-0.3cm}
    \caption{\small \textbf{Offline model-based optimization} (MBO) requires generating designs $\x$ that optimize a black-box objective function $f(\x)$ using a given static dataset of designs, without any active queries to the ground truth function.}
    \label{fig:offline_mbo_figure}
    \vspace{-0.2cm}
\end{figure}
Although online black-box optimization has been studied extensively, offline MBO has received comparatively less attention, and only a small number of recent works study offline MBO in the setting with high-dimensional design spaces~\cite{brookes2019conditioning, kumar2019model, fannjiang2020autofocused,fu2021offline,trabuccoconservative}.
This is partly because online techniques cannot be directly applied in settings where offline MBO is used, especially in high-dimensional settings. Online techniques, such as Bayesian optimization~\cite{snoek2012practical}, often require iterative feedback via queries to the objective function. Such online optimizers exhibit optimistic behavior: they rely on active queries at completely unseen designs irrespective of whether such a design is good or not. When access to these queries is removed, certain considerations change: optimism is no longer desirable and distribution shift becomes a major challenge~\cite{kumar2019model}.

Even with only a few existing offline MBO methods, it is hard to compare and track progress, as methods are typically proposed and evaluated on different tasks with distinct evaluation protocols. To the best of our knowledge, there is no commonly adopted benchmark for offline MBO. To address, we introduce a suite of tasks for offline MBO with a standardized evaluation protocol. We include a diverse set of tasks that span a wide range of application domains---from synthetic biology to robotics--that aims at representing the core challenges in real-world offline MBO.
While the tasks are not intended to directly enable solving the corresponding real-world problems, which would require a lot of machinery in real hardware setup (e.g., a real robot or access to a wetlab for molecule design), they are intended to provide algorithm designers with a representative sampling of challenges that reflect the difficulties with real-world MBO. That is to say, the tasks are not intended to be \emph{real}, but are intended to be \emph{realistically challenging}.
Further, the diversity of the tasks measures how they generalize across multiple domains and verifies they are not specialized to a single task.
Our benchmark incorporates a variety of challenging factors, such as high dimensionality and sensitive discontinuous objective functions, which help identify the strengths and weaknesses of MBO methods.
Along with this benchmark suite, we present reference implementations of a number of existing offline MBO and baseline optimization methods. We systematically evaluate them on all of the proposed benchmark tasks and report results.
We hope that our work can provide insight into the progress of offline MBO methods and serve as a meaningful metric to galvanize research in this area.

\section{Offline Model-Based Optimization (Offline MBO) Problem Statement} 
\label{sec:problem_statement}
In online model-based optimization, the goal is to optimize a (possibly stochastic) black-box objective function $\oracle(\design)$ with respect to its input. The objective can be written as $\arg\max_{\design}\oracle(\design)$. Methods for online MBO typically optimize the objective iteratively, proposing design $\design_k$ at the $k$th iteration and query the objective function to obtain $\oracle(\design_k)$.
Unlike its online counterpart, access to the true objective $\oracle$ is not available in offline MBO. Instead, the algorithm $\algo$ is provided access to a static dataset $\dataset = \{(\design_i, \score_i)\}$ of designs $\design_i$ and a corresponding measurement of the objective value $\score_i$. The algorithm consumes this dataset and produces an optimized candidate design $\optima$ which is evaluated against the true objective function. This paradigm is illustrated in Figure~\ref{fig:offline_mbo_figure}. Abstractly, the objective for offline MBO is:
\vspace{-5pt}
\begin{align}
    \arg\max_{\algo} \oracle(\optima) \ \text{where} \ \optima = \algo(\dataset).
\end{align}
\vspace{5pt}
In practice, producing a single optimal design entirely from offline data is very difficult, so offline MBO methods are more commonly evaluated~\citep{kumar2019model}
in terms of ``$P$ percentile of top $K$'' performance, where the algorithm produces $K$ candidates and the $P$ percentile objective value determines final performance. Next we discuss two important aspects pertaining to offline MBO, namely, why offline MBO algorithms can improve beyond the best design observed in the offline dataset despite no active queries, and the associated challenges with devising offline model-based optimization algorithms.  

\begin{figure}[h]
\centering
\vspace{-7pt}
\includegraphics[width=0.49\linewidth]{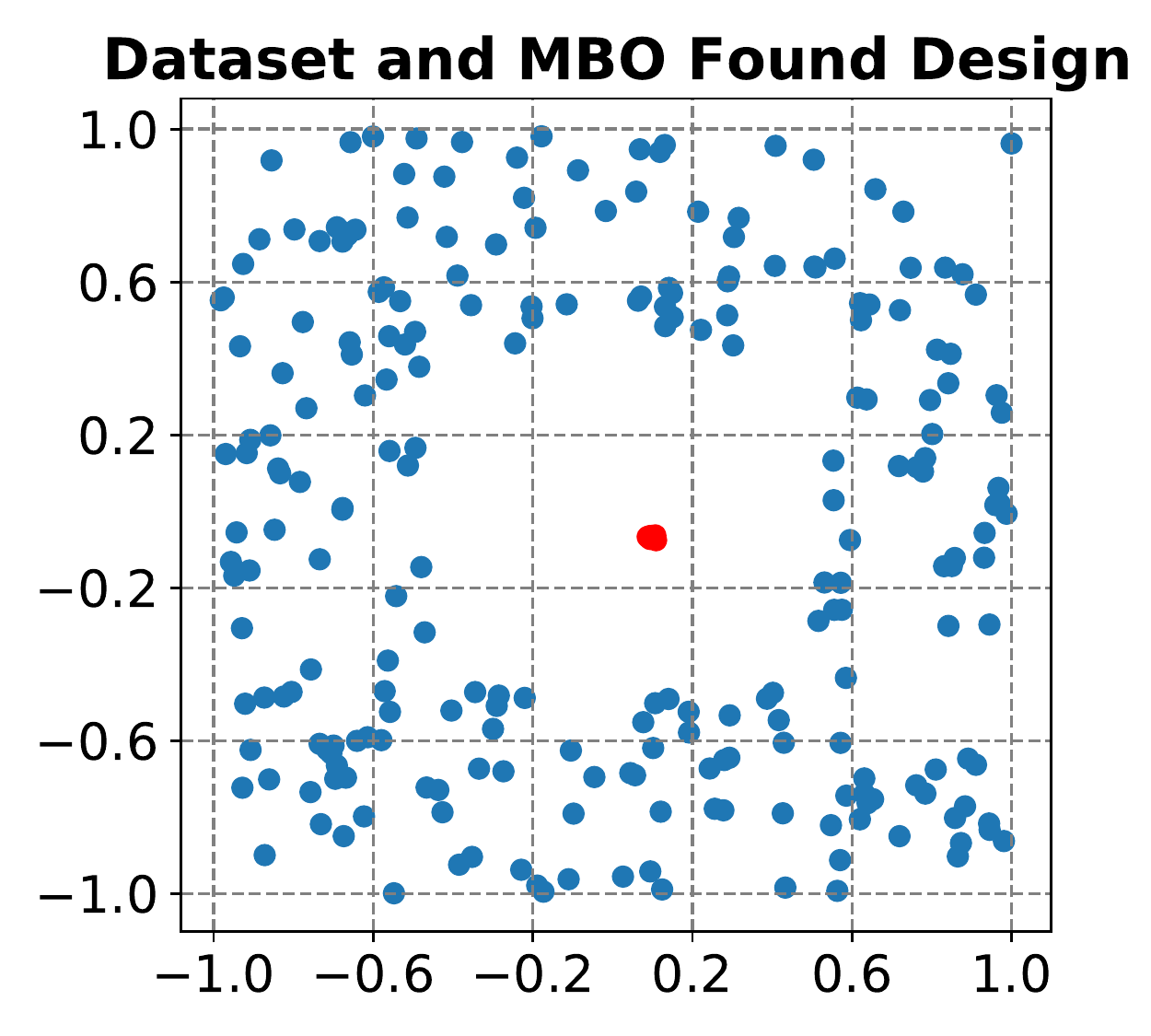}
\includegraphics[width=0.49\linewidth]{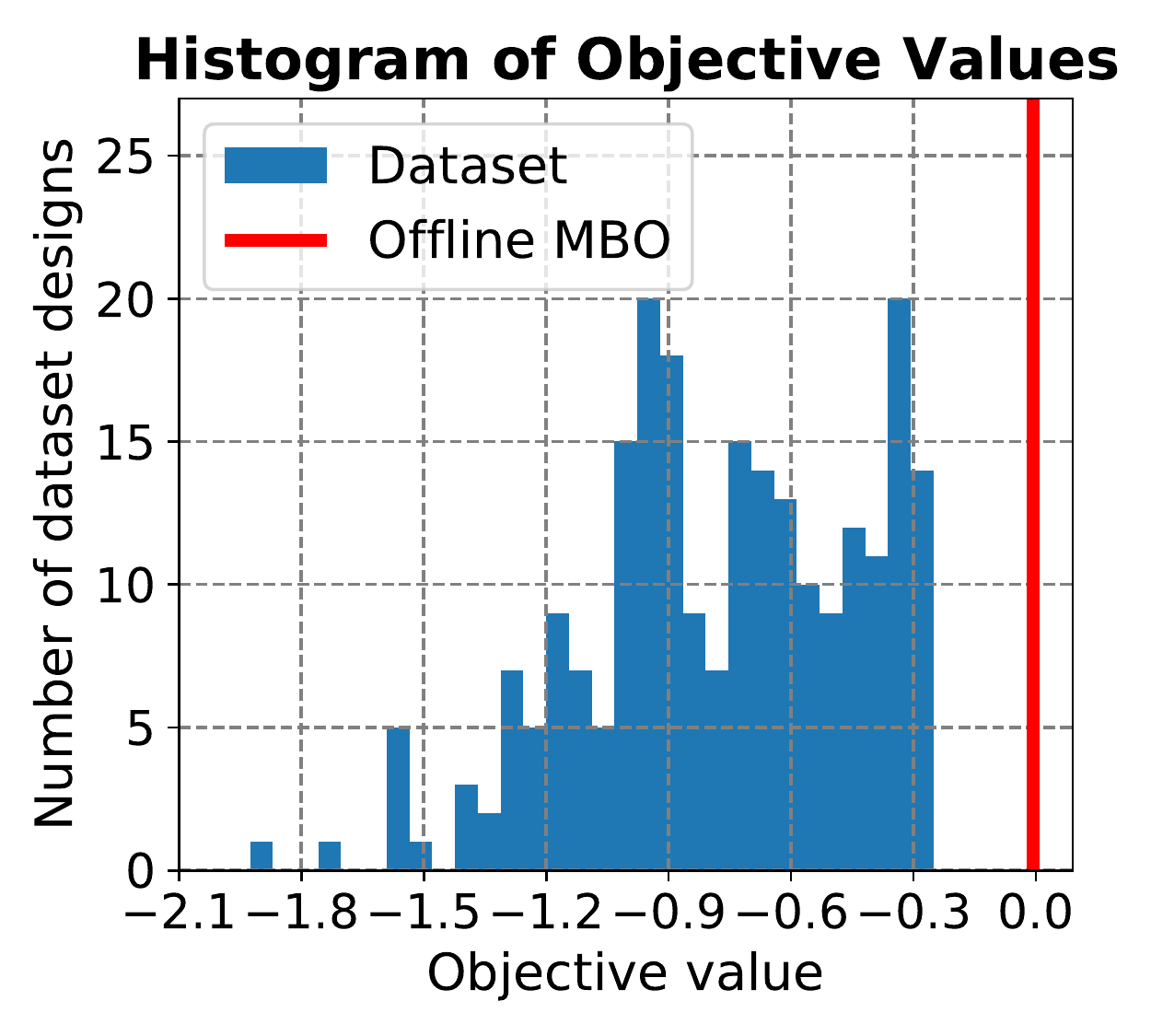}
\vspace{-0.3cm}
\caption{\label{fig:toy_task} \footnotesize{Offline MBO finds designs better than the best in the observed dataset by exploiting compositional structure of the objective function. \textbf{Left:} datapoints in a toy quadratic function MBO task over 2D space with optimum at $(0.0, 0.0)$ in blue, MBO found design in red. \textbf{Right:} Objective value for optimal design is much higher than that observed in the dataset.}}
\vspace{-5pt}
\end{figure}

\textbf{Would offline MBO even produce designs better than the best observed design in the dataset?}\\
A natural question to ask is whether it is even reasonable to expect offline MBO algorithms to improve over the performance of the best design seen in the dataset. As we will show in our benchmark results, many of the tasks that we propose do already admit solutions from existing algorithms that exceed the performance of the best sample in the dataset. To provide some intuition for how this can be possible, consider a simple example of offline MBO problems, where the objective function $f(\design)$ can be represented as a sum of functions of independent partitions of the design variables, i.e., $f(\design) = f_1(\design[1]) + f_2(\design[2]) + \cdots + f_N(\design[N]))$, where $\design[1], \cdots, \design[N]$ denotes disjoint subsets of design variables $\design$. The dataset of the offline MBO problem contains optimal design variable for each partition, but not the combination. If an offline MBO algorithm can identify the compositional structure of independent partitions, it would be able to combine the optimal design variable for each partition together to form the overall optimal design and therefore improving the performance over the best design in the dataset.
To better demonstrate this idea, we created a toy problem in two dimensions, where the objective function is simply $f(x, y) = -x^2 - y^2$.   We then run a na\"ive gradient ascent algorithm, as we will describe later in this paper. In Figure~\ref{fig:toy_task}, we can clearly see that our offline MBO algorithm is able to learn to combine the best $x$ and $y$ and produce designs significantly better than the best sample in the dataset.
Such a condition appears in a number of scenarios in practice e.g., in reinforcement learning (RL), where the Markov structure provides a natural decomposition satisfying this composition criterion~\citep{fu2020d4rl} and effective offline RL algorithms are known to exploit this structure~\citep{fu2020d4rl}  or in protein design, where objective such as fluorescence naturally decompose into functions of neighboring amino acids~\citep{brookes2019conditioning}.

\textbf{What makes offline MBO especially challenging?}

The offline nature of the problem prevents the algorithm $\algo$ from querying the ground truth objective $\oracle$ with its proposed design candidates,
and this makes the offline MBO problem much more difficult than the online design optimization problem. One na\"{i}ve approach to tackle this problem is to learn a model of the objective function using the dataset, which we can denote $\hat{f}(\design)$, and then convert this offline MBO problem into a regular online MBO problem
by treating the learned objective model as the true objective.  However, this generally does not work: optimizing the design $\design$ with respect to a learned proxy $\hat{f}(\design)$ will produce \emph{out-of-distribution} designs that ``fool'' $\hat{f}(\design)$ into outputting a high value, analogously to adversarial examples. Indeed, it is well known that optimizing na\"{i}vely with respect to model inputs to obtain a desired output will usually simply ``fool'' the model~\citep{kumar2019model}.
A na\"{i}ve strategy to address this out-of-distribution issue is to constrain the design to stay close to the data, but this is also problematic, since in order to produce a design that is better than the best training point, it is usually necessary to deviate from the training data at least somewhat. In almost all practical MBO problems, such as optimization over drug molecules or robot morphologies as we discuss in section~\ref{sec:challenges}, designs with the highest objective values typically lie on the tail of the dataset distribution and we may not find them by staying extremely close to the data distribution. 
This conflict between the need to remain close to the data to avoid out-of-distribution inputs and the need to deviate from the data to produce better designs is one of the core challenges of offline MBO. This challenge is often exacerbated in real-world settings by the high dimensionality of the design space and the sparsity of the available data, as we will show in our benchmark. A good offline MBO method needs to carefully balance these two sides, producing optimized designs that are good, but not too far from the data distribution.

\begin{table*}[t!]
    \centering
    \begin{tabular}{l||r|r|r|r|r}
        \toprule
        \textbf{Dataset Name} & \textbf{Size} & \textbf{Dimensions} & \textbf{Categories} & \textbf{Type} & \textbf{Oracle} \\
        \midrule
        \textbf{TF Bind 8} & 32898 & 8 & 4 & Discrete & Exact \\
        \textbf{TF Bind 10} & 50000 & 10 & 4 & Discrete & Exact \\
        \textbf{NAS} & 1771 & 64 & 5 & Discrete & Exact \\
        \textbf{ChEMBL} & 1093 & 31 & 591 & Discrete & Random Forest \\

        \midrule
        \textbf{Superconductor} & 21263 & 86 & N/A & Continuous & Random Forest \\
        \textbf{Ant Morphology} & 25009 & 60 & N/A & Continuous & Exact \\
        \textbf{D'Kitty Morphology} & 25009 & 56 & N/A & Continuous & Exact \\
        \textbf{Hopper Controller} & 3200 & 5126 & N/A & Continuous & Exact \\
        \bottomrule
    \end{tabular}
    \caption{\small{\textbf{Overview of the tasks in our benchmark suite.} Design-Bench includes a variety of tasks from different domains, including several from prior work, and multiple new tasks, with both discrete and continuous design spaces, making it suitable for benchmarking offline MBO methods. In addition to the provided tasks, we explore several from prior work in Appendix~\ref{sec:unused_tasks} that we chose not to include in the final benchmark.}}
    \label{tab:task_overview}
    \vspace{-15pt}
\end{table*}

\section{Related Work}
\label{sec:related}
Prior work has extensively focused on online or active MBO, which requries active querying on the ground truth function, including algorithms using Bayesian optimization and their scalable variants~\citep{lizotte2008practical, snoek2012practical,snoek2015scalable,shahriari2015taking,perrone2017multiple}, direct search~\citep{kolda2003optimization},
genetic or evolutionary algorithms~\citep{whitley1994genetic,pal2012comparative,yang2020firefly}, the cross-entropy method~\citep{rubinstein2013cross}, simulated annealing~\citep{van1987simulated}, etc. These methods may not be well suited for real-world problems where the ground truth function is expensive to evaluate and therefore prohibitive for active querying.
Offline MBO utilizes an already existing database of designs and objective values, which might be obtained from previously conducted experiments. This presents an attractive algorithmic paradigm towards approaching such scenarios. Since offline MBO prohibits any ability to query the true objective with new designs, it presents different challenges from those typically studied in online MBO problem, as we discuss in Section~\ref{sec:challenges}. These new challenges in turn require new benchmarks, motivating our work.

The most important components for a good offline MBO benchmark are datasets that capture the challenges of real-world problems. Fortunately, researchers working on a wide variety of scientific fields have already collected many datasets of designs which we can use for training offline MBO algorithms.
ChEMBL \cite{Gaulton2012ChEMBL} provides a dataset for drug discovery, where molecule activities are measured against a target assay. \citet{hamidieh2018superconductor} analyze the critical temperatures for superconductors and provide a dataset to search for room-temperature superconductors with potential in the construction of quantum computers. Some of these datasets have already been employed in the study of offline MBO methods~\cite{kumar2019model, brookes2019conditioning, fannjiang2020autofocused}.
However, these studies all use different sets of tasks and their evaluation protocols are highly domain-specific, making it difficult to compare across methods. In our benchmark, we incorporate modified variants of some of these datasets along with our own tasks, and provide a standardized evaluation protocol.

Recently, several methods have been proposed for specifically addressing the offline MBO problem. These methods~\citep{kumar2019model,brookes2019conditioning,fannjiang2020autofocused} typically learn models of the objective function and optionally, a generative model~\citep{kingma2013auto,goodfellow2014generative,mirza2014conditional} of the design manifold and use them for optimization. We discuss these methods in detail in Section~\ref{sec:algorithms} and benchmark their performance in Section~\ref{sec:exps}.

\section{Design-Bench Benchmark Tasks}
\label{sec:tasks}

In this section, we describe the set of tasks included in our benchmark. An 
overview of the tasks is provided in Table~\ref{tab:task_overview}.
Each task in our benchmark suite comes with a dataset $\dataset = \{(\design_i, \score_i)\}$, along with a ground-truth oracle objective function $\oracle$ that can be used for evaluation. An offline MBO algorithm should not query the ground-truth oracle function during training, even for hyperparameter tuning. We first discuss the nature of oracles used in Design-Bench.

\textbf{Expert model as oracle function.} While in some of the tasks in our benchmark, such as tasks pertaining to robotics (D'Kitty Morphology, and Ant Morphology), the oracle functions are evaluated by running computer simulations to obtain the true objective values, in the other tasks, the true objective values can only be obtained by conducting expensive physical experiments. While the eventual aim of offline MBO is to make it possible to optimize designs in precisely such settings, requiring real physical experiments for evaluation makes the design and benchmarking of new algorithms difficult and time consuming. Therefore, to facilitate benchmarking, we follow the evaluation methodology in prior work~\cite{brookes2019conditioning, fannjiang2020autofocused} and use models built by domain experts as our ground-truth oracle functions. Note, however, that the training data provided for offline MBO is still real data -- the domain expert model is used \emph{only} to evaluate the result for benchmarking purposes. In many cases, these expert models are \emph{also} learned, but with representations that are hand-designed, with built-in domain-specific inductive biases. The ground-truth oracle models are also trained on much more data than is made available for solving the offline MBO problem, which increases the likelihood that this expert model can provide an accurate evaluation of solutions found by offline MBO, even if they lie outside the training distribution. While this approach to evaluation diminishes the realism of our benchmark since these proxy ``true functions'' may not always be accurate, we believe that this trade off is worthwhile to make benchmarking practical. The main purpose of our benchmark is to facilitate the evaluation and development of offline MBO algorithms, and we believe that it is important to include tasks in domains where the true objective values can only be obtained via physical experiments, which make up a large portion of the real-world MBO problems.

We now provide a detailed description of the tasks in our benchmark. A description of the data collection strategy and pre-processing can be found in Appendix~\ref{app:task_descriptions}.

\textbf{Superconductor: critical temperature maximization.}
The Superconductor task is taken from the domain of materials science, where the goal is to design the chemical formula for a superconducting material that has a high critical temperature. We adapt a real-world dataset proposed by~\citet{hamidieh2018superconductor}. The dataset contains 21263 superconductors annotated with critical temperatures. Prior work has employed this dataset for the study of offline MBO methods~\cite{fannjiang2020autofocused}, and we follow their convention using a random forest regression model, detailed in \cite{hamidieh2018superconductor}, for our oracle. The model achieves a final Spearman's rank-correlation coefficient with a held-out validation set of 0.9210. The design space for Superconductor is a vector with 86 real-valued components representing the mixture of elements by number of atoms in the chemical formula of each superconductor.

\begin{wrapfigure}{r}{0.5\columnwidth}
\vspace{-0.5cm}
\centering
\includegraphics[width=0.98\linewidth]{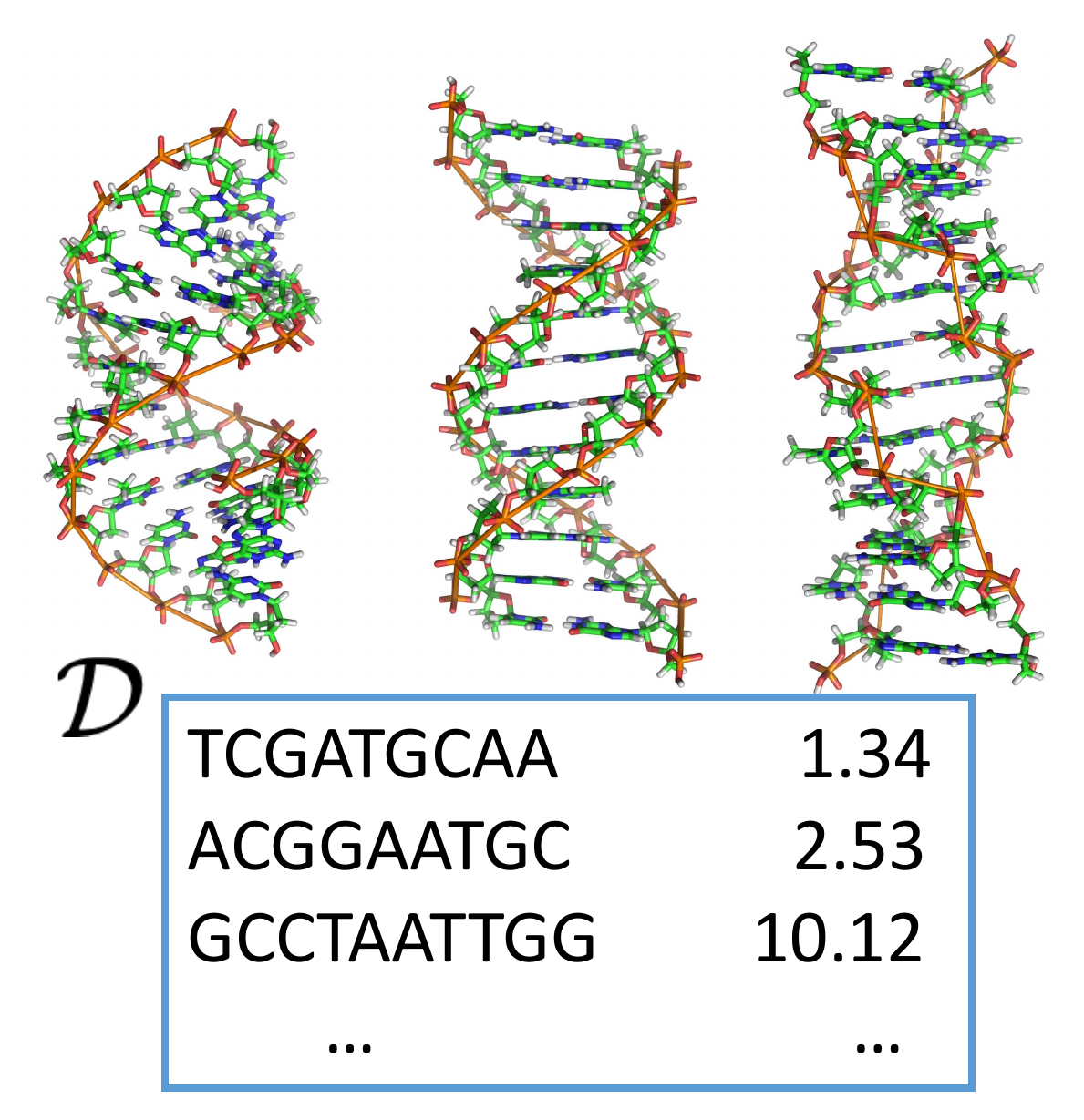}
\vspace{-0.7cm}
\end{wrapfigure}
\textbf{TF Bind 8 and TF Bind 10: DNA sequence optimization.}
The goal of TF Bind 8 and TF Bind 10 is to find the length-8 DNA sequence with maximum binding affinity with a particular transcription factor (\texttt{SIX6\_REF\_R1} by default). The ground truth binding affinities for all 65,792 and 1,048,576 designs for the two tasks are available~\citep{barrera2016survey}. The design space consists of sequences of one of four categorical variables, one for each nucleotide. For TF Bind 8, we sample 32898 of all the sequences, and for TF Bind 10 we sample 50000 sequences to form the training set.

\begin{wrapfigure}[7]{R}{0.39\columnwidth}
\vspace{-15pt}
\centering
\includegraphics[width=0.4\columnwidth]{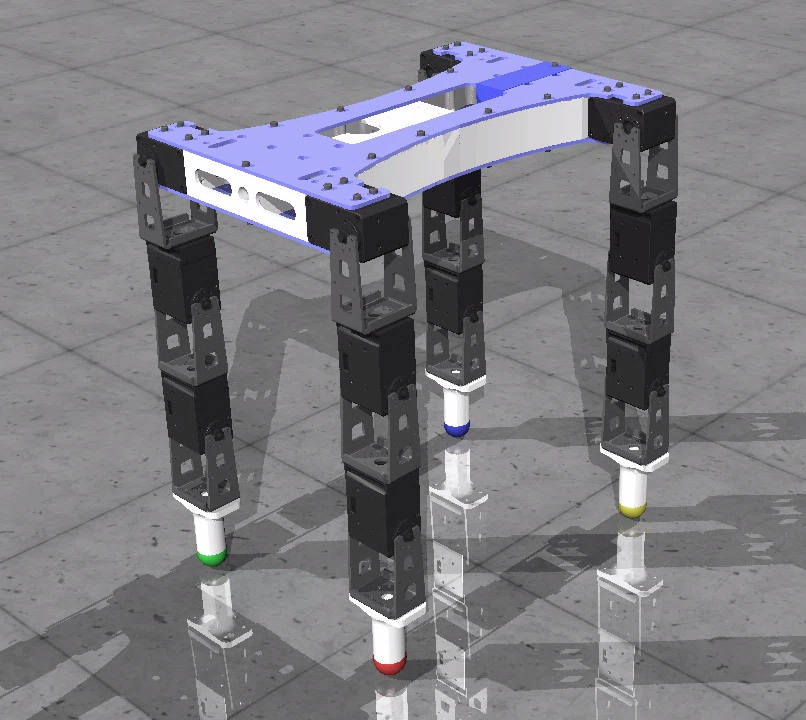}
\end{wrapfigure}
\textbf{Ant and D'Kitty Morphology: robot morphology optimization.} The goal is to optimize the morphological structure of two simulated robots: Ant from OpenAI Gym~\cite{brockman2016openai} and D'Kitty from ROBEL~\cite{Kumar_ROBEL}. For Ant Morphology, the we need to optimize the morphology of a quadruped
robot to run as fast as possible. For D'Kitty Morphology, the goal is to optimize the morphology of D'Kitty robot (shown on the right) to navigate the robot to a fixed location. Thus the goal is to find robot morphologies optimal for the given tasks. In order to control the robot with the generated morphology, we use a controller that has been optimized for the given morphology with the Soft Actor Critic algorithm~\cite{haarnoja2018soft}.
The morphology parameters of both robots include size, orientation, and location of the limbs, giving us 60 continuous values in total for Ant and 56 for D'Kitty.
To evaluate a given design, we run robotic simulation in the MuJoCo~\cite{todorov2012mujoco} simulator for 100 time steps, averaging 16 independent trials giving us reliable but cheap to compute estimates. 

\begin{wrapfigure}{r}{0.46\columnwidth}
\vspace{-0.7cm}
\centering
\hspace{-0.2cm} \includegraphics[width=1.1\linewidth]{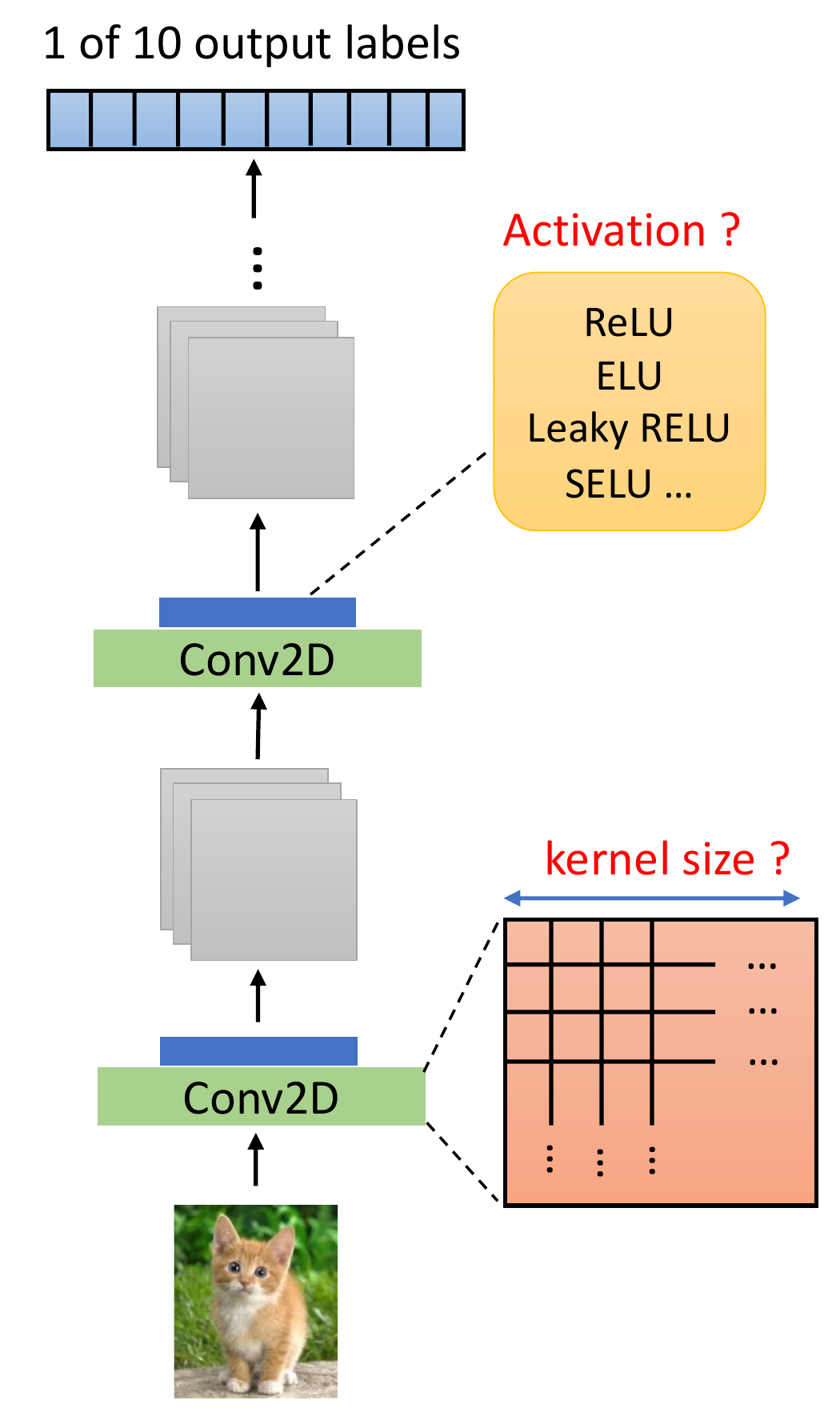}
\vspace{-0.7cm}
\end{wrapfigure}
\textbf{NAS: neural architecture search on CIFAR10.}
The goal of this task is to search for a good neural network architecture~\cite{zoph2016neural} to optimize the test accuracy on the CIFAR10~\cite{hinton2012improving} dataset. The model is a 32-layer convolutional neural network with residual connections, and the task requires searching over the kernel sizes and activation function types for each of the 32 layers. Given the small image size of CIFAR10, we choose the list of possible kernel sizes to be $\{2, 3, 4, 5, 6\}$. The possible choices of activation functions are ReLU, ELU, leaky ReLU, SELU~\cite{klambauer2017self} and SiLU~\cite{elfwing2018sigmoid}. The combination of kernel sizes and activation functions give us a 64 dimensional discrete space with 5 categories per dimension. The dataset is collected by randomly sampling architectures in the search space. We evaluate the design by training the produced architecture on the training CIFAR10 dataset for 20 epochs and evaluating the accuracy on the test set.

\begin{wrapfigure}{r}{0.52\columnwidth}
\vspace{-0.5cm}
\centering
\includegraphics[width=\linewidth]{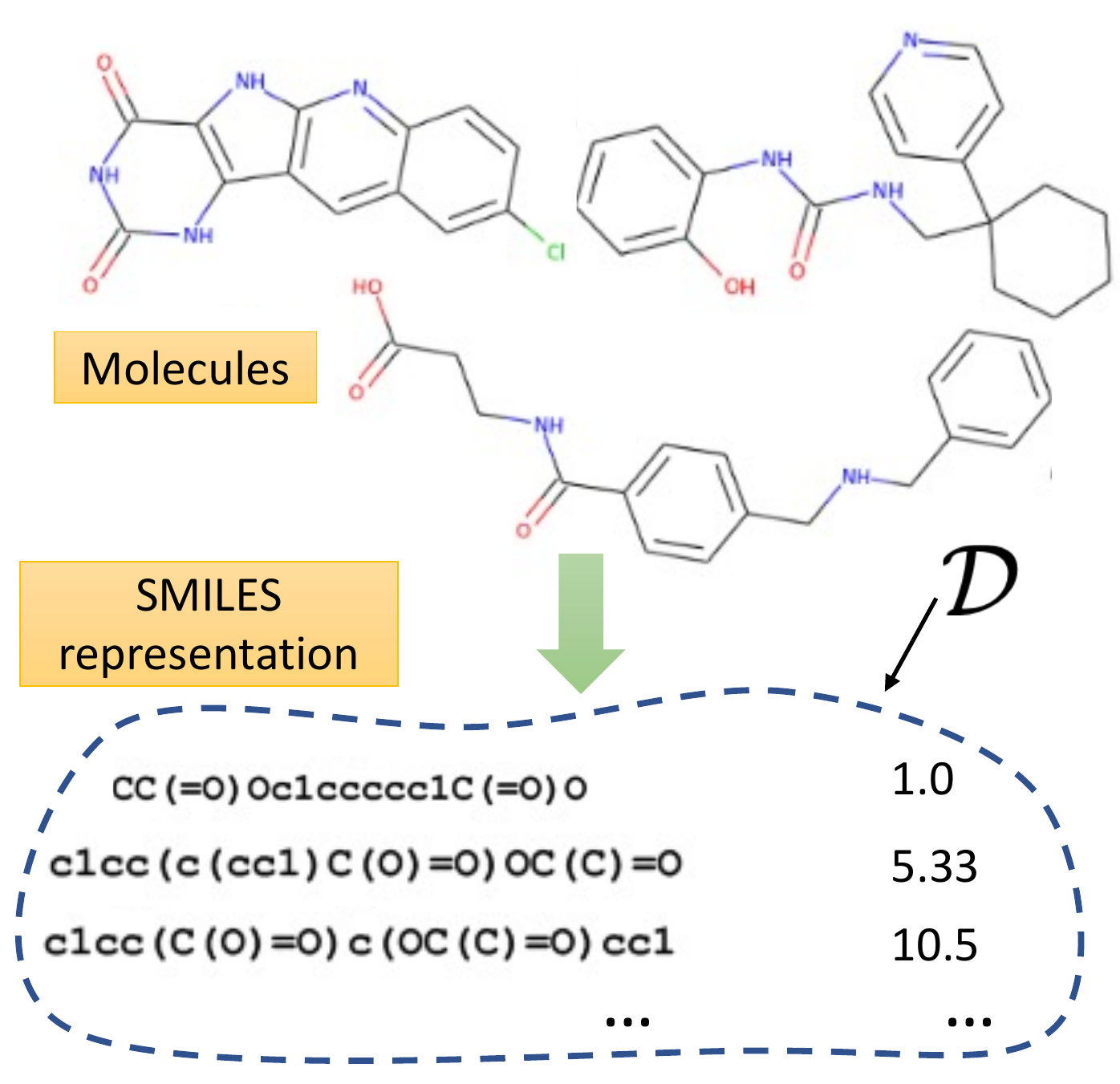}
\vspace{-1.0cm}
\end{wrapfigure}
\textbf{ChEMBL: molecule activity maximization for drug discovery.}
The ChEMBL task in Design bench is derived from a large-scale drug property database from which the task name is derived \cite{Gaulton2012ChEMBL}. This database consists of pairs of molecules and assays tested for a particular chemical properties. We choose the assay whose ChEMBL id is \texttt{CHEMBL3885882} and measure its \texttt{MCHC} value. The goal of the resulting optimization problem is to design a molecule that, when paired with assay \texttt{CHEMBL3885882}, achieves a high \texttt{MCHC} value. The training set is restricted to molecules whose SMILES \cite{weininger1988smiles} encoding has fewer than 30 tokens. This results in a training set with 1093 samples, and a design space of length 31 sequences of categorical variables that take one of 591 values.

\textbf{Hopper Controller: robot neural network controller optimization.}
The goal in this task is to optimize the weights of a neural network policy so as to maximize the expected discounted return on the Hopper-v2 locomotion task in OpenAI Gym~\cite{brockman2016openai}. While this might appear similar to reinforcement learning (RL), our formulation is distinct: unlike RL, we don't have access to any form of trajectory data in the dataset. 
Instead, our dataset only comprises of neural network controller weights and the corresponding return values, which invalidates the applicability of conventional RL methods. We evaluate the true objective value of any design by running 1000 steps of simulation in the MuJoCo simulator conventionally ussed with this environment. 
The design space of this task is high-dimensional with 5126 continuous variables corresponding to the flattened weights of a neural network controller. The dataset is collected by training a PPO~\cite{Schulman2017ppo} and recording the agent's weights every 10,000 samples.

\section{Task Properties, Challenges and Considerations}
\label{sec:challenges}
The primary goal of our benchmark is to provide a general test bench for developing, evaluating, and comparing algorithms for offline MBO. While in principle any online active black-box optimization problem can be turned into an offline MBO problem by collecting a dataset of designs and corresponding objective measurements, it is important to pick a subset of tasks that represent the challenges of real-world problems in order to convincingly evaluate algorithms and obtain insights about algorithm behavior. Therefore, several factors must be considered when choosing the tasks, which we discuss next.

\textbf{Diversity and realistically challenging.} First of all, the tasks need to be diverse and realistically challenging in order to prevent offline MBO algorithms from overfitting to a particular problem domain and to expect that methods performing well on this benchmark suite would also perform well on real-world offline MBO problems. Design-Bench consists of tasks that are diverse in many respects. It includes both tasks with \textit{discrete} and with \textit{continuous} design spaces. Continuous design spaces, equipped with metric space and ordering structures, could make the problem easier to solve than discrete design spaces. However, discrete design spaces are finite and therefore might enjoy better dataset coverage than some continuous tasks. While our tasks are not intended to directly solve real-world problems (e.g., we don't actually expect the best robot morphology in our benchmark to actually correspond to the best possible real robot morphology due to a variety of factors including limitations of the simulator), they are intended to provide designers with a representative sampling of challenges that reflect the kinds of difficulties they would face with real-world datasets, making them realistically challenging.

\begin{figure*}[h!]
    \centering
    \vspace{-4pt}
    \includegraphics[trim=150 0 150 0,clip,width=0.9\linewidth]{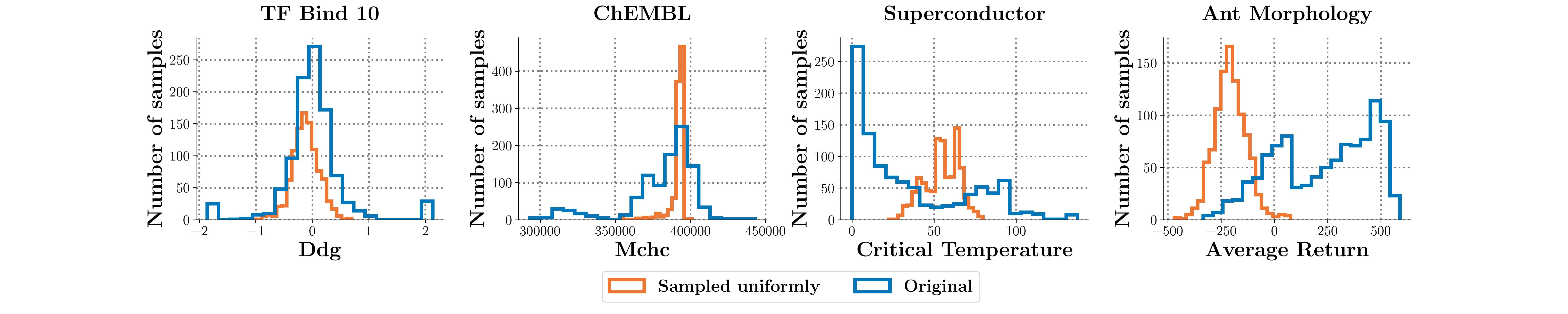}
    \vspace{-8pt}
    \caption{\small{\textbf{Histogram (frequency distribution) of objective values in the dataset compared to a uniform re-sampling of the dataset} from the design space. In every case, re-sampling skews the distribution of values to the left, suggesting that there exists a thin manifold of valid designs in the high-dimensional design space, and most of the volume in this space is occupied by low-scoring designs. The distribution of objective values in the dataset are often heavy-tailed, for instance, in the case of ChEMBL and Superconductor.}}
    \label{fig:manifold}
    \vspace{-8pt}
\end{figure*}

\textbf{High-dimensional design spaces.} In many real-world offline MBO problems, such as drug discovery~\cite{Gaulton2012ChEMBL}, the design space is \textit{high-dimensional} and good designs sparsely lie on a \textit{thin manifold} in this high-dimensional space. This poses a challenge for many MBO methods: to be effective on such problem domains, MBO methods need to capture the thin manifold to be able to produce good designs.
Prior work~\cite{kumar2019model} has noted that this can be very hard in practice. In our benchmark, we include a task derived from ChEMBL with up $31$ dimensions and $591$ categories per dimension to capture this challenge.
To intuitively understand this challenge, we performed a study on some tasks in Figure~\ref{fig:manifold}, where we sampled 3200 designs uniformly at random from the design space and plotted a histogram of the objective values against those in the dataset we provide, which only consists of valid designs. Observe the discrepancy in objective values, where randomly sampled designs generally attain objective values lower than the best dataset sample. This suggests that performant designs only lie on a thin manifold in the design space and therefore we are very unlikely to hit a performant design by uninformed random sampling.

\begin{figure}[h!]
    \centering
    \includegraphics[width=0.7\linewidth]{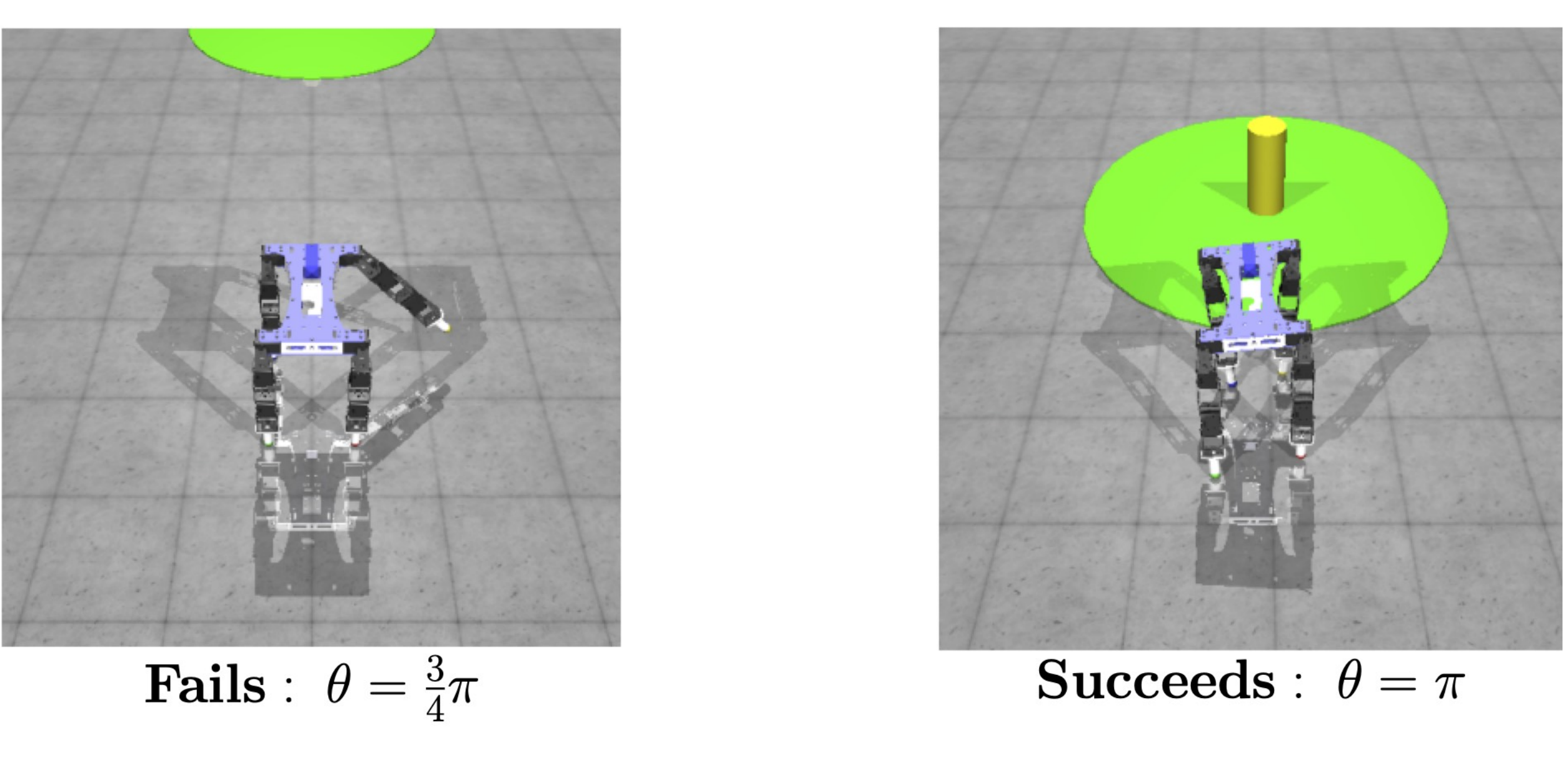}
    \includegraphics[trim=500 0 500 35,clip,width=0.9\linewidth]{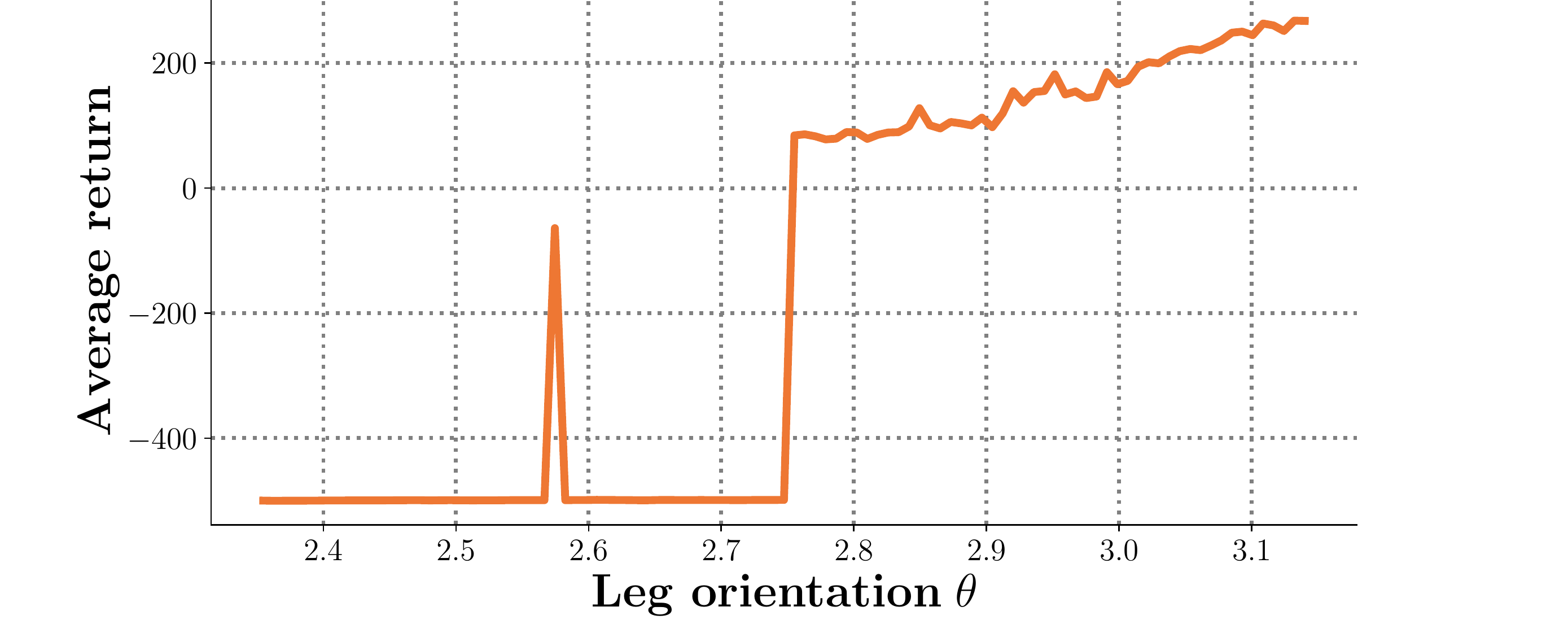}
    \vspace{-8pt}
    \caption{\small{\textbf{Highly sensitive landscape of the ground truth objective function in DKittyMorphology.} A small change in a single dimension of the design space, for instance changing the orientation $\theta$ (x-axis) of the base of the robot's front right leg, critically impacts the performance value (y-axis). The robot's
    design on the left is the original D'Kitty design and is held constant while varying $\theta$ uniformly from $\frac{3}{4}\pi$ to $\pi$.}}
    \label{fig:sensitivity}
    \vspace{-16pt}
\end{figure}

\textbf{Highly sensitive objective function.} Another important challenge that should be taken into consideration is the \textit{high sensitivity} of objective functions, where closeness of two designs in design space need not correspond to closeness in their objective values, which may differ drastically. This challenge is naturally present in practical problems like drug discovery~\cite{Gaulton2012ChEMBL}, where the change of a single atom could significantly alter the property of the molecule. The DKitty Morphology and Ant Morphology tasks in our benchmark suite are also particularly challenging in this respect. To visualize the high sensitivity of the objective function, we plot a one dimensional slice of the objective function around a single sample in our dataset in Figure~\ref{fig:sensitivity}. Observe that with other variables kept constant, slightly altering one variable can significantly reduce the objective value, making it hard for offline MBO methods to produce the optimal design.

\textbf{Heavy-tailed data distributions.} Finally, another challenging property for offline MBO methods is the shape of the data distribution. Learning algorithms are likely to exhibit poor learning behavior when the distribution of objective values in the dataset is heavy-tailed. This challenge is often present in black-box optimization~\citep{ChowdhuryG19} and can hurt the performance of MBO algorithms that use a generative model as well as those that use a learned model of the objective function. As shown in Figure~\ref{fig:manifold} tasks in our benchmark exhibit this heavy-tailed structure.

\section{Algorithm Implementations}
\label{sec:algorithms}
To provide a baseline for comparisons in future work, we benchmark a number of recently proposed offline MBO algorithms on each of our tasks. Since some of our tasks have a high input dimensionality, we chose prior methods that can handle \emph{both} the case of offline training data (i.e., no active interaction) and high-dimensional inputs. Thus, we include MINs~\cite{kumar2019model}, CbAS~\cite{brookes2019conditioning}, autofocusing CbAS~\cite{fannjiang2020autofocused} and REINFORCE/CMA-ES~\citep{Williams92} in our comparisons, along with a baseline na\"{i}ve ``gradient ascent'' method that approximates the true function $f(\bx)$ with a deep neural network and then performs gradient ascent on the output of this model. In this section, we briefly discuss these algorithms, before performing a comparative evaluation in the next section. Our implementation of these algorithms are open sourced and can be found at \href{https://github.com/rail-berkeley/design-baselines}{github.com/rail-berkeley/design-baselines}.

\textbf{Gradient ascent (Grad).} This is a simple baseline that learns a model of the objective function, $\hat{f}(\bx)$, and optimizes $\bx$ against this learned model via gradient ascent. Formally, the optimal solution $\bx^*$ generated by this method can be computed as a fixed point of the following update: $\bx_{t+1} \leftarrow \bx_t + \alpha \nabla_\bx \hat{f}(\bx)\vert_{\bx = \bx_{t}}$.
In practice we perform $T=200$ gradient steps, and report $\bx_T$ as the final solution. Such methods are susceptible to producing invalid solutions, since the learned model does not capture the manifold of valid-designs and hence cannot constrain the resulting $\bx_T$ to be on the manifold. We additionally evaluate a variant (\textbf{Grad. Min}) optimizing over the minimum prediction of $N=5$ learned objective functions in an ensemble of learned objective functions and (\textbf{Grad. Mean}) that optimizes the mean ensemble prediction. We discuss additional tricks (e.g., normalization of inputs and outputs) that we found beneficial with this baseline in Appendix~\ref{sec:grad_normalization}.

\textbf{Covariance matrix adaptation (CMA-ES).} CMA-ES \citet{Hansen06} is a simple optimization algorithm that maintains a belief distribution over the optimal design, and gradually refines this distribution by adapting the covariance matrix using feedback from a (learned) objective function, $\hat{f}(\bx)$. Formally, let $\bx_t \sim \mathcal{N}(\mu_t, \Sigma_t)$ be the samples obtained from the distribution at an iteration $t$, then CMA-ES computes the value of learned $\hat{f}(\bx_t)$ on samples $\bx_t$, and fits $\Sigma_{t+1}$ to the highest scoring fraction of these samples and repeats this multiple times.
The learned $\hat{f}(\bx)$ is trained via supervised regression.

\textbf{REINFORCE~\citep{Williams92}.} We also evaluated a method that optimizes a learned objective function, $\hat{f}(\bx)$, using the REINFORCE-style policy-gradient estimator. REINFORCE is capable of handling non-smooth and highly stochastic objectives, making it an effective choice. This method parameterizes a distribution $\pi_\theta(\bx)$ over the design space and then updates the parameters $\theta$ of this distribution towards the design that maximizes $\hat{f}(\bx)$, using the gradient, $\mathbb{E}_{\bx \sim \pi_\theta(\bx)}[\nabla_\theta \log \pi_\theta(\bx) \cdot \hat{f}(\bx)]$.
We train an ensemble of $\hat{f}(\bx)$ models and pick the subset of models that satisfy a validation loss threshold $\tau$. This threshold is task-specific; for example, $\tau \leq 0.25$ is sufficient for Superconductor-v0.

\textbf{Conditioning by adaptive sampling (CbAS)~\citep{brookes2019conditioning}.} CbAS learns a density model in the space of design inputs, $p_0(\bx)$ that approximates the data distribution and gradually adapts it towards the optimized solution $\bx^*$. In a particular iteration $t$, CbAS alternates between \textbf{(1)} training a variational auto-encoder (VAE)~\cite{kingma2013auto} on a set of samples generated from the previous model $\dataset_{t} = \{\bx_i\}_{i=1}^m; \bx_i \sim p_{t-1}(\cdot)$ using a weighted version of the standard ELBO objective biased towards \textit{estimated} better designs and \textbf{(2)} generating new design samples from the autoencoder to serve as $\dataset_{t+1} = \{\bx_i | \bx_i \sim p_t(\cdot) \}$. In order to estimate the objective values for designs sampled from the learned density model $p_t(\bx)$, CbAS utilizes separately trained models of the objective function, $\hat{f}(\bx)$ trained via supervised regression. This training process, at a given iteration $t$, is:
\vspace{-8pt}
\begin{align}
    p_{t+1}(\bx) & := \arg \min_{p} \frac{1}{m} \sum_{i=1}^{m} \frac{p_{0}(\bx_i)}{p_t(\bx_i)} P ( \hat{f}(\bx_i) \geq \tau ) \log p_t(\bx_i) \nonumber \\
    & ~~\text{where}~~ \{\bx_i\}_{i=1}^m \sim p_t(\cdot). 
\label{eqn:cbas}
\end{align}

\vspace{-10pt}
\textbf{Autofocused CbAS (Auto. CbAS)~\citep{fannjiang2020autofocused}.}~~Since CbAS uses a learned model of the objective function $\hat{f}(\bx)$ to iteratively adapt the generative model $p(\bx)$ towards the optimized design, the function$\hat{f}(\bx)$ will inevitably be required to make predictions on shifting design distributions $p_t(\bx)$. Hence, any inaccuracy in these values can adversely affect the optimization procedure. Autofocused CbAS aims to correct for this  shift by re-training $\hat{f}(\bx)$ (now denoted $\hat{f}_t(\bx)$) under the design distribution given by the current model, $p_t(\bx)$ via importance sampling, which is then fed into CbAS.
\vspace{-5pt}
\begin{equation*}
    \hat{f}_{t+1} := \arg \min_{\hat{f}}~~ \frac{1}{|\mathcal{D}|} \sum_{i=1}^{|\mathcal{D}|} \frac{p_t(\bx_i)}{p_{0}(\bx_i)} \cdot \left( \hat{f}(\bx_i) - y_i \right)^2,
\end{equation*}
\textbf{Model inversion networks (MINs)~\citep{kumar2019model}.} MINs learn an inverse map from the objective value to a design, $\hat{f}^{-1}: \mathcal{Y} \rightarrow \mathcal{X}$ by using objective-conditioned inverse maps, search for optimal $y$ values during optimization and finally query the learned inverse map to produce the corresponding optimal design. MIN minimizes a divergence measure $\gL_p(\dataset) := \E_{y \sim p_\dataset(y)}\left[ D(p_\dataset(\bx|y), \hat{f}^{-1}(\bx|y)) \right]$ to train such an inverse map. During optimization, MINs obtains the optimized design by sampling from the inverse map conditioned on the optimal $y$-value.

\textbf{Bayesian optimization (BO-qEI).} We perform offline Bayesian optimization to maximize the value of a learned objective function $\hat{f}(\bx)$ by fitting a Gaussian Process, proposing candidate solutions, and labeling these candidates using $\hat{f}(\bx)$. To improve efficiency, we use the quasi-Expected-Improvement acquisition function \cite{reparameterization2017} based on the BoTorch framework \cite{botorch2019}. 

\textbf{Conservative Objective Models (COMs)~\citep{trabuccoconservative}}. COMs utilizes a single learned model of the objective function $\hat{f}(\bx_i)$ for offline model-based optimization. COMs learns a conservative model of the objective function using an augmented regression objective that penalizes overestimation of the performance on off-manifold designs $\bx$. Solutions are obtained by initializing $\bx_0$ to a design from an observed training set $\dataset$, and performing $T$ steps of gradient ascent $\bx_{t+1} \leftarrow \bx_{t} + \nabla_{\bx} \alpha \hat{f}(\bx) |_{\bx = \bx_{t}} $ on the conservative objective model's predictions with respect to the design $\bx$.

\begin{figure*}[ht]
    \centering
    \vspace{-4pt}
    \includegraphics[width=0.8\linewidth]{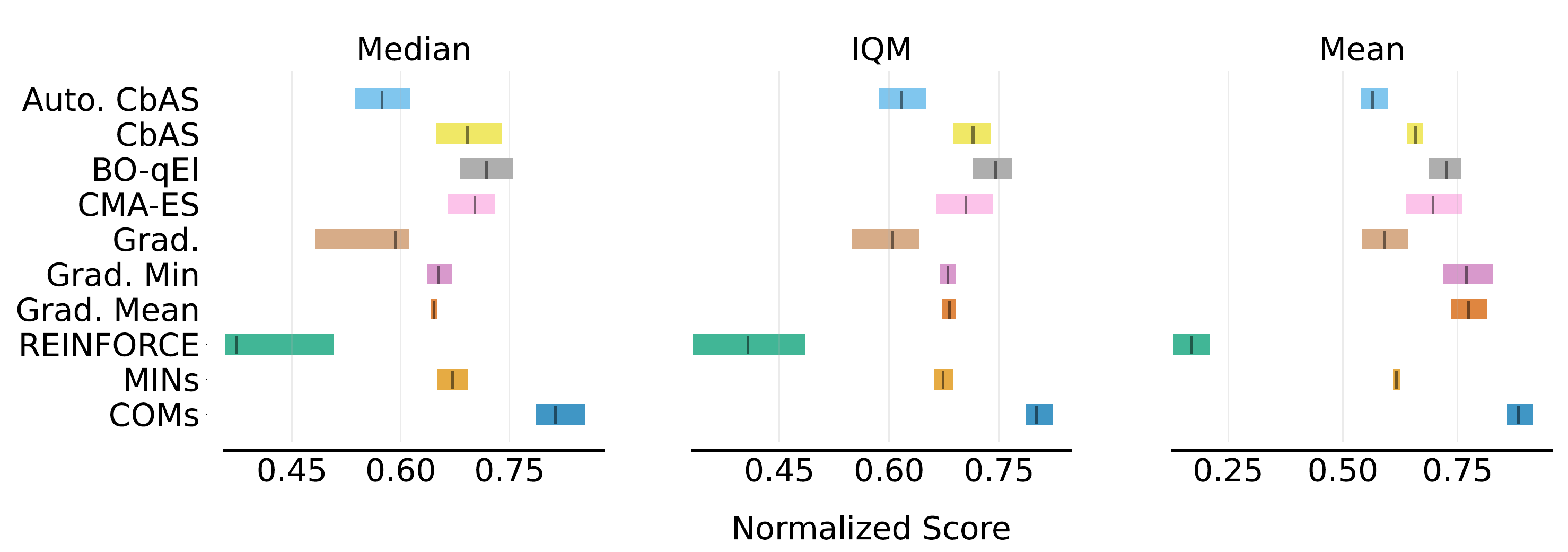}
    \vspace{-8pt}
    \caption{\small \textbf{Median, IQM and mean}~\citep{agarwal2021deep} aggregated 100th percentile normalized scores (with 95\% Stratified Bootstrap CIs) for the tasks in Design-Bench.}
    \label{fig:interval}
    \vspace{-5pt}
\end{figure*}

\begin{table*}[ht]
\begin{scriptsize}
    \centering
    \begin{tabular}{l||r|r|r|r|r|r|r|r}
    \toprule
    {} & \textbf{TF Bind 8} & \textbf{TF Bind 10} & \textbf{ChEMBL} & \textbf{NAS} & \textbf{Superconductor} & \textbf{Ant Morph.} & \textbf{DKitty Morph.} & \textbf{Hopper}\\
    \midrule
    $\mathcal{D}$ (\textbf{best}) &  0.439         &  0.467         &  0.605          &   0.436   &  0.400         &  0.565         &  0.884 & 1.0 \\
   \textbf{ Auto. CbAS}              &  0.910 ± 0.044 &  0.630 ± 0.045 &  0.249 ± 0.305 &   0.506 ± 0.074 &  0.421 ± 0.045 &  0.882 ± 0.045 &  0.906 ± 0.006 & 0.137 ± 0.005\\
    \textbf{CbAS}                          &  0.927 ± 0.051 &  0.651 ± 0.060 &  0.473 ± 0.264 &   0.683 ± 0.079 &  0.503 ± 0.069 &  0.876 ± 0.031 &  0.892 ± 0.008 &  0.141 ± 0.012  \\
   \textbf{ BO-qEI  }                      &  0.798 ± 0.083 &  0.652 ± 0.038 &  0.596 ± 0.226 &   1.079 ± 0.059 &  0.402 ± 0.034 &  0.819 ± 0.000 &  0.896 ± 0.000 & 0.550 ± 0.118 \\
    \textbf{CMA-ES}                        &  0.953 ± 0.022 &  0.670 ± 0.023 &  0.085 ± 0.225 &   0.985 ± 0.079 &  0.465 ± 0.024 &  1.214 ± 0.732 &  0.724 ± 0.001 & 0.604 ± 0.215\\
    \textbf{Grad. }              &  0.977 ± 0.025 &  0.657 ± 0.039 &  0.307 ± 0.308 &   0.433 ± 0.000  &  0.518 ± 0.024 &  0.293 ± 0.023 &  0.874 ± 0.022 & 1.035 ± 0.482 \\
    \textbf{Grad. Min}  &  0.984 ± 0.012 &  0.649 ± 0.032 &  0.653 ± 0.024 &   0.433 ± 0.000 &  0.506 ± 0.009 &  0.479 ± 0.064 &  0.889 ± 0.011 & 1.391 ± 0.589 \\
    \textbf{Grad. Mean} &  0.986 ± 0.012 &  0.645 ± 0.018 &  0.652 ± 0.005 &   0.433 ± 0.000 &  0.499 ± 0.017 &  0.445 ± 0.080 &  0.892 ± 0.011 & 1.586 ± 0.454 \\
    \textbf{REINFORCE }                    &  0.948 ± 0.028 &  0.663 ± 0.034 &  0.164 ± 0.285 &  -1.895 ± 0.000 &  0.481 ± 0.013 &  0.266 ± 0.032 &  0.562 ± 0.196 & -0.020 ± 0.067\\
    \textbf{ MINs}                          &  0.905 ± 0.052 &  0.616 ± 0.021 &  0.000 ± 0.000 &   0.717 ± 0.046  &  0.499 ± 0.017 &  0.445 ± 0.080 &  0.892 ± 0.011 & 0.424 ± 0.166\\
    \textbf{COMs  }                        &  0.973 ± 0.016 &  0.730 ± 0.136 &  0.633 ± 0.000 &   0.459 ± 0.139 &  0.439 ± 0.033 &  0.944 ± 0.016 &  0.949 ± 0.015 & 2.056 ± 0.314 \\
    \bottomrule
    \end{tabular}
    \vspace{-5pt}
    \caption{\textbf{100th percentile} evaluations. Results are averaged over 8 trials, and $\pm$ indicates the standard deviation of the performance. The objective value normalization procedure is described in Appendix~\ref{sec:score_normalization}.}
    \label{tab:perf100}
    \vspace{-5pt}
\end{scriptsize}
\end{table*}

\section{Benchmarking Prior Methods}
\label{sec:exps}
In this section, we provide a comparison of prior algorithms discussed in Section~\ref{sec:algorithms} on our proposed tasks. For purposes of standardization, easy benchmarking, and future algorithm development, we present results for all Design-Bench tasks in Table~\ref{tab:perf100}. As discussed in Section~\ref{sec:problem_statement}, we allow each method to produce $K = 128$ optimized design candidates. These candidates are then evaluated with the oracle function, and we report the $100^{\text{th}}$ percentile performance among them averaged over 8 independent runs, following the conventions of prior works~\citep{fannjiang2020autofocused, brookes2019conditioning,kumar2019model}. We also provide unnormalized and $50^\text{th}$\%ile results in Appendices~\ref{sec:unnorm_scores}, \ref{sec:perf50}.

\textbf{Algorithm setup and hyperparameter tuning.}
Since our goal is to generate high-performing solutions without \emph{any} knowledge of the ground truth function, any form of hyperparameter tuning on the parameters of the learned model should crucially respect this evaluation boundary and tuning must be performed completely offline, agnostic of the objective function. We provide a recommended method for tuning each algorithm described in Section~\ref{sec:algorithms} in Appendix~\ref{sec:hparam_select}, which also serves as a set of guidelines for tuning future methods with similar components. 

To briefly summarize, \textbf{for \textbf{CbAS}}, hyperparameter tuning amounts to tuning a VAE where samples from the prior distribution map to on-manifold designs after reconstruction.
We empirically found that a $\beta$-VAE was essential for stability of CbAS and high values of $\beta > 1$ are especially important for modelling high-dimensional spaces.
As a general task-agnostic principle for selecting $\beta$, we choose the smallest $\beta$ such that the VAE's latent space does not collapse during importance sampling. Collapsing latent-spaces seem to coincide with diverging importance sampling, and the VAE's reconstructions collapsing to a single mode. \textbf{For \textbf{MINs}}, hyperparameter tuning amounts to fitting a good generative model. We observe that MINs is particularly sensitive to the scale of $y_{i}$ when conditioning, which we resolve by normalizing the objective values. We implement MINs using WGAN-GP, and find that similar hyperparameters work well across domains. \textbf{For \textbf{Gradient Ascent}}, while prior works report poor performance for na\"ive gradient ascent optimization on top of learned models of the objective function, we find that by normalizing the designs $\bx$ and objective values $y$ to have unit normal statistics and scaling the learning rate as $\alpha \leftarrow \alpha \sqrt{d}$ where $d$ is the dimension of the design space (discussed in Appendix~\ref{sec:grad_normalization}), a na\"ive gradient ascent based procedure performs reasonably well on most tasks without task-specific tuning. For discrete tasks, only the objective values are normalized, and optimization is performed over log-probabilities of designs. We then obtain optimized designs by running 200 steps of gradient ascent starting from the top scoring $128$ samples in each dataset. We provide further details in Appendix~\ref{sec:hparam_select}.

\textbf{Results.}
The results for all tasks are provided in Table~\ref{tab:perf100}. There are several takeaways from these results.
First, these results confirm that three prior offline MBO methods (MINs, CbAS, and Autofocused CbAS), are very successful at solving a wide range of offline MBO problems of varying dimensional and modality. Furthermore, perhaps somewhat surprisingly, a classical CMA-ES baseline is competitive with several highly sophisticated MBO methods in 4 out of 8 tasks (Table~\ref{tab:perf100}). This result suggests that it might be difficult for generative models to capture high-dimensional task distributions with enough precision to be used for optimization, and in a number of tasks, these components might be unnecessary. Additionally a naive gradient ascent baseline is competitive with complex approaches utilizing generative modelling on 4 of the 8 tasks. However, on the other hand, as described in Appendix~\ref{sec:grad_normalization} and \ref{sec:tuning_gradient_ascent},  baseline is also sensitive to certain design choices such as input normalization schemes and the number of optimization steps $T$. Therefore, while not a full-fledged offline MBO method, we believe that gradient ascent has potential to form a fundamental building block for future offline MBO methods.

Finally, we remark that the performance of methods in Table~\ref{tab:perf100} differ from the those reported by prior works. This difference stems from the standardization procedure employed in dataset generation (which we discuss in Appendix~\ref{app:task_descriptions}).

\section{Discussion and Conclusion}\label{sec:conclusion}
Offline model-based optimization carries the promise to convert existing databases of designs into powerful optimizers, without the need for expensive real-world experiments for actively querying the ground truth objective function. However, due to the lack of standardized benchmarks and evaluation protocols, it has been difficult to accurately track the progress of offline MBO methods. To address this problem, we introduce Design-Bench, a benchmark suite of offline MBO tasks that covers a wide variety of domains, and both continuous and discrete, low and high dimensional design spaces. We provide a comprehensive evaluation of existing methods under identical assumptions. The comparatively high efficacy of even simple baselines such as CMA-ES and na\"ive gradient ascent suggests the need for careful tuning and standardization of methods in this area. 
An interesting avenue for future work in offline MBO is to devise methods that can be used to perform model and hyperparameter selection. One promising approach to address this problem is to devise methods for offline evaluation of produced solutions. We hope that our benchmark will be adopted as the standard metric in evaluating offline MBO algorithms and provides insight in future algorithm development. 

\vspace{-5pt}

\section*{Acknowledgements}
We thank members of the RAIL lab at UC Berkeley, Jennifer Listgarten and Clara Fannjiang for informative discussions and suggestions for tasks appearing in this benchmark. We thank anonymous reviewers from NeurIPS and ICLR for feedback on a previous version of this manuscript. This research is supported by Intel, Schmidt Futures, the Office of Naval Research and compute resources from Google Cloud and Microsoft Azure.


\bibliography{references}
\bibliographystyle{icml2022}

\clearpage
\begin{appendices}

\section{Data Collection}
\label{app:task_descriptions}

In this section, we detail the data collection steps used for creating each of the tasks in design-bench. We answer \textbf{(1) }where is the data from, and \textbf{(2)} what pre-processing steps are used?

\subsection{TF Bind 8 and TF Bind 10}

The TF Bind 8 and TF Bind 10 tasks are derivativesof the transcription factor binding activity survey performed by \cite{barrera2016survey}, where the binding activity scores of every possible DNA sequence was measured with a variety of human transcription factors. We filter the dataset by selecting a particular transcription factor \texttt{SIX6\_REF\_R1}, and defining an optimization problem where the goal is to synthesize a length 8 DNA sequence with high binding activity with human transcription factor \texttt{SIX6\_REF\_R1}. This particular transcription factor for TF Bind 8 was recently used for optimization in \cite{angermueller2019model, angermuller2020p3b0}. TF Bind 8 is a fully characterized dataset containing 65792 samples, representing every possible length 8 combination of nucleotides $\bx_{\text{TFBind8}}\in\{0,1\}^{8\times4}$. The training set given to offline MBO algorithms is restricted to the bottom 50\%, which results in a visible training set of 32898 samples.

\subsection{ChEMBL}

The ChEMBL task is a derivative of a much larger dataset that is derived from ChEMBL \cite{Gaulton2012ChEMBL}, a large database of chemicals and their properties. The datawas originally collected by performing physical experiments on a large number of molecules, and measuring a chemical property in the presence of a target assay. We have processed the ChEMBL database---available at \url{https://www.ebi.ac.uk/chembl/g/#browse/activities}---into collections of smaller datasets mapping particular molecules to measured values, determined by a target assay that accompanies each set. We choose the assay specified by \texttt{ASSAY\_CHEMBL\_ID} = \texttt{CHEMBL3885882} and select the standard type of \texttt{MCHC} as the measurement to maximize with offline model-based optimization. The resulting dataset has 1093 samples in total. This assay is chosen for its high validation rank correlation, namely 0.7141, when fitting a random forest regression model to map molecules to \texttt{MCHC} values. The majority of other assays in ChEMBL produce a validation rank correlation below 0.5. We preprocess the dataset by converting each molecule into a SMILES string using RDKit, and then apply the DeepChem \texttt{SmilesTokenizer} to convert each SMILES string into a sequence of integer tokens. We then remove all molecules whose SLIMES sequence is longer than a maximum of 31 tokens with the vocabulary has 591 elements, $\bx_{\text{ChEMBL}}\in\{0,1\}^{31\times591}$. When evaluating MBO methods, we remove the top 50\% of molecules sorted by their \texttt{MCHC} value to increase task difficulty.

\subsection{Superconductor}

The Superconductor task is inspired by recent work \cite{fannjiang2020autofocused} that applies offline MBO to optimize the properties of superconducting materials for high critical temperature. The data we provide in our benchmark is real-world superconductivity data originally collected by \cite{hamidieh2018superconductor}, and subsequently made available to the public at \url{https://archive.ics.uci.edu/ml/datasets/Superconductivty+Data#}. The original dataset consists of superconductors featurized into vectors containing measured physically properties like the number of chemical elements present, or the mean atomic mass of such elements. One issue with the original dataset that was used in \cite{fannjiang2020autofocused} is that the numerical representation of the superconducting materials did not lend itself to recovering a physically realizable material that could be synthesized in a lab after performing model-based optimization. In order to create an \textit{invertible} input specification, we deviate from prior work and encode superconductors as vectors whose components represent the number of atoms of specific chemical elements present in the superconducting material---a serialization of the chemical formula of each superconductor. The result is a real-valued design space with 86 components $\bx_{\text{Superconductor}}\in\mathcal{R}^{86}$. The full dataset used to learn approximate oracles for evaluating MBO methods has 21263 samples, but we restrict this number to 17010 (the 80th percentile) for the training set of offline MBO methods to increase difficulty. 

\subsection{Ant \& D'Kitty Morphology}

Both morphology tasks are collected by us, and share methodology. The goal of these tasks is to design the morphology of a quadrupedal robot---an ant or a D'Kitty---such that the agent is able to crawl quickly in a particular direction. In order to collect data for this environment, we create variants of the MuJoCo Ant and the ROBEL D'Kitty agents that have parametric morphologies. The goal is to determine a mapping from the morphology of the agent to the average return of the agent using a controller optimized for that morphology. In order to facilitate fast optimization, we pre-compute a morphology conditioned neural network controller using SAC \cite{Haarnoja2018SoftAO} that has been trained to perform optimally on a wide range of morphologies. For both the Ant and the D'Kitty, we train the controllers for more than ten million environment steps, and a maximum episode length of 200, with all other settings as default. These morphology conditioned controllers are trained on Gaussian distributions of morphologies. The Gaussian distributions are obtained by adding Gaussian noise with standard deviation 0.03 for Ant and 0.01 for D'Kitty the design-space range to the default morphologies. After obtaining trained morphology-conditioned controllers, we create a dataset of morphologies for model-based optimization by sampling initialization points randomly, and then using CMA-ES to optimize for morphologies that attain high reward using the morphology-conditioned controllers. To obtain initialization points, we add Gaussian random noise to the default morphology for the Ant with standard deviation 0.075 and D'Kitty with standard deviation 0.1, and then apply CMA-ES with standard deviation 0.02. We ran CMA-ES for 250 iterations and then restarted, until 25000 morphologies were collected, resulting in 25009 samples for both the Ant and D'Kitty. The design space for Ant morphologies is $\bx_{\text{Ant}}\in\mathcal{R}^{60}$, whereas for D'Kitty morphologies is $\bx_{\text{D'Kitty}}\in\mathcal{R}^{56}$. We sremove the top 40\% of samples when training offline MBO algorithms.

\subsection{NAS}
The data for the NAS task is collected by us. The goal of this task is to search for a good neural network architecture to optimize the test accuracy on the CIFAR10 dataset. The architecture search space is a 64-dimensional discrete variable with 5 categories for each dimension. We collect the dataset by randomly sample architecture designs in the search space, and train them on the CIFAR10 dataset. We sample 2440 total designs, and select the bottom performing 70\% to be our training set. This gives us 1771 samples in total, with the test accuracy ranging from 59.3\% to 63.8\%. 

\subsection{Hopper Controller}
The goal of this task is to design a set of weights for as neural network policy, in order to achieve high expected return when evaluating that policy. The data collected for Hopper Controller was taken by training a three layer neural network policy with 64 hidden units and 5126 total weights on the Hopper-v2 MuJoCo task using Proximal Policy Optimization \cite{Schulman2017ppo}. Specifically, we use the default parameters for PPO provided in stable baselines \cite{stable-baselines}. The dataset we provide with this benchmark has 3200 unique weights. In order to collect this many, we run 32 experimental trials of PPO, where we train for one million steps, and save the weights of the policy every 10,000 environment steps. The policy weights are represented originally as a list of tensors. We first traverse this list and flatten each of the tensors, and we then concatenate each of these flattened tensors into a single training example $x_{\text{Hopper}}\in\mathcal{R}^{5126}$. The result is an optimization problem over neural network weights. After collecting these weights, we perform no additional pre-processing steps. In order to collect objective score values we perform a single rollout for each $x$ using the Hopper-v2 MuJoCo environment. The horizon length for training and evaluation is limited to 1000 simulation time steps. 

\section{Oracle Functions}

We detail oracle functions for evaluating ground truth scores for each of the tasks in design-bench. A common thread among these is that the oracle, if trained, is fit to a larger static dataset containing higher performing designs than observed by a downstream MBO algorithm.

\subsection{TF Bind 8 and TF Bind 10}
\vspace{-0.1cm}
TF Bind 8 and TF Bind 10 are a fully characterized discrete offline MBO tasks, which means that all possible designs have been evaluated \cite{barrera2016survey} and are contained in the full hidden datasets. The oracles are therefore implemented simply as a lookup table that returns the score corresponding to a particular DNA sequence from the dataset. By restricting the size of the training set visible to an offline MBO algorithm, it is possible for the algorithm to propose a design that achieves a higher score than any other DNA sequence visible to the algorithm during training.

\subsection{ChEMBL}
\vspace{-0.1cm}
We tested several models as candidate oracle functions for ChEMBL \cite{Gaulton2012ChEMBL}, including a Gaussian Process, Random Forest, CNN, and Transformer regression models. We ultimately chose the Random Forest model in scikit-learn due to its quick inference and relatively high performance compared with neural network alternatives, achieving a spearman's rank correlation coefficient of 0.7141 with a held-out validation set. These models were trained on the entire hidden ChEMBL dataset for \texttt{ASSAY\_CHEMBL\_ID} = \texttt{CHEMBL3885882} with standard type \texttt{MCHC} encoded into SMILES and tokenized. Hyper-parameters for the random forest oracle are provided in the official github release of design-bench.

\subsection{Superconductor}
\vspace{-0.1cm}
The Superconductor oracle function is also a random forest regression model. The model we use it the model described by \cite{hamidieh2018superconductor}. We borrow the hyperparameters described by them, and we use the RandomForestRegressor provided in scikit-learn. Similar to the setup for the previous set of tasks, this oracle is trained on the entire hidden dataset of superconductors. The random forest has a rank correlation of 0.9155 with a held-out validation set.

\subsection{Ant \& D'Kitty Morphology}
\vspace{-0.1cm}
The Ant \& D'Kitty Morphology tasks in design-bench use an exact oracle function, using the MuJoCo simulator. For both morphology tasks, the simulator performs a rollout and returns the sum of rewards at every timestep in that rollout. Each task is accompanied by a pre-trained morphology-conditioned policy. To perform evaluation, a morphology is passed to the Ant or D'Kitty MuJoCo environments respectively, and a dynamic-morphology agent is initialized inside these environments. These simulations can be time consuming to run, and so we limit the rollout length to 100 steps. The morphology conditioned policies were trained using the reinforcement learning algorithm SAC for 10 million steps for each task, and are ReLU networks with two hidden layers of size 64.

\subsection{NAS}
The NAS task in the design bench uses an exact oracle, where we train the proposed architecture on CIFAR10 and then test it on the test set. To perform the evaluation, we construct the proposed architecture using PyTorch, and train it for 20 epochs using batch size 256 and then compute the test accuracy on the test set.

\subsection{Hopper Controller}
Unlike the previously described tasks, Hopper Controller implements an exact oracle function. For Hopper Controller the oracle takes the form of a single rollout using the Hopper-v2 MuJoCo environment. The designs for Hopper Controller are neural network weights, and during evaluation, a policy with those weights is instantiated---in this case that policy is a three layer neural network with 11 input units, two layers with 64 hidden units, and a final layer with 3 output units. The intermediate activations between layers are hyperbolic tangents. After building a policy, the Hopper-v2 environment is reset and the reward for 1000 time-steps is summed. That summed reward constitutes the score returned by the Hopper Controller oracle. The limit of performance is the maximum return that an agent can achieve in Hopper-v2 over 1000 steps.

\section{Experimental Details}
\label{sec:exp_details}
In this section we present additional details for the experiments, including the score normalization process and 50th percentile performance.

\subsection{Objective Normalization}\label{sec:score_normalization}
In order to report performance on the same order of magnitude for each offline model-based optimization task in Design-Bench, we normalize the performance reported in Table~\ref{tab:perf100} by calculating the minimum objective value $y_{\min}$ and the the maximum objective value $y_{\max}$ in the full unobserved dataset associated with each offline model-based optimization problem. \emph{Crucially}, note that this is not the same as normalizing with respect to the best and worst samples in the training dataset used by the offline MBO algorithm, but rather a bigger dataset of designs and objective values. We then report performance by calculating what fraction of the distance between $y_{\min}$ and $y_{\max}$ is attained by a particular offline MBO baseline.
\begin{equation}
    y_{\text{normalized}} (y) = \frac{y - y_{\min}}{y_{\max} - y_{\min}}
\end{equation}
The final performance $y_{\text{normalized}}$ is the normalized performance of an offline MBO method that achieved an unprocessed objective value of $y$. The result is larger than one when the offline MBO method finds a solution more performance than all solutions in the full unobserved dataset associated with the corresponding task. The result is less than zero when the offline MBO method finds a solution attaining less performance than all samples in the full unobserved dataset. 

\subsection{50th Percentile Experiment Results}\label{sec:perf50}
In this section, we present the 50th percentile performance of the runs presented in main paper in Table~\ref{tab:perf100}. Similar to the 100th percentile performance reported in the main text, performance is calculated by evaluating solutions to each task found by an optimization method, subtracting the minimum objective value present in the corresponding task dataset, and dividing by the range of objective values present in the corresponding task dataset. The result is a performance of greater than one if optimization converges to a solution with a higher objective value that the best observed design in the corresponding task dataset.

\begin{table*}[h]
\begin{scriptsize}
    \centering
    \begin{tabular}{l||r|r|r|r|r|r|r}
    \toprule
    {} &      TF Bind 8 &     TF Bind 10 &         ChEMBL &       NAS & Superconductor &            Ant Morphology &         D'Kitty Morphology \\
    \midrule
    Auto. CbAS                    &  0.419 ± 0.007 &  0.461 ± 0.007 &  -1.823 ± 0.000 &   0.217 ± 0.005 &  0.131 ± 0.010 &   0.364 ± 0.014 &  0.736 ± 0.025 \\
    CbAS                          &  0.428 ± 0.010 &  0.463 ± 0.007 &  -1.807 ± 0.004 &   0.292 ± 0.027 &  0.111 ± 0.017 &   0.384 ± 0.016 &  0.753 ± 0.008 \\
    BO-qEI                        &  0.439 ± 0.000 &  0.467 ± 0.000 &  -1.774 ± 0.020 &   0.544 ± 0.099 &  0.300 ± 0.015 &   0.567 ± 0.000 &  0.883 ± 0.000 \\
    CMA-ES                        &  0.537 ± 0.014 &  0.484 ± 0.014 &  -1.763 ± 0.019 &   0.591 ± 0.102 &  0.379 ± 0.003 &  -0.045 ± 0.004 &  0.684 ± 0.016 \\
    Gradient Ascent               &  0.609 ± 0.019 &  0.474 ± 0.005 &  -1.772 ± 0.018 &   0.433 ± 0.000 &  0.476 ± 0.022 &   0.134 ± 0.018 &  0.509 ± 0.200 \\
    Grad. Min                     &  0.645 ± 0.030 &  0.470 ± 0.002 &  -1.769 ± 0.014 &   0.433 ± 0.000 &  0.471 ± 0.016 &   0.185 ± 0.008 &  0.746 ± 0.034 \\
    Grad. Mean                    &  0.616 ± 0.023 &  0.471 ± 0.004 &  -1.757 ± 0.010 &   0.433 ± 0.000  &  0.469 ± 0.022 &   0.187 ± 0.009 &  0.748 ± 0.024 \\
    MINs                          &  0.421 ± 0.015 &  0.468 ± 0.006 &  -1.745 ± 0.000 &   0.433 ± 0.000 &  0.336 ± 0.016 &   0.618 ± 0.040 &  0.887 ± 0.004 \\
    REINFORCE                     &  0.462 ± 0.021 &  0.475 ± 0.008 &  -1.805 ± 0.003 &  -1.895 ± 0.000 &  0.463 ± 0.016 &   0.138 ± 0.032 &  0.356 ± 0.131 \\
    COMs                          &  0.497 ± 0.038 &  0.465 ± 0.008 &   0.633 ± 0.000 &   0.287 ± 0.173 &  0.386 ± 0.018 &   0.519 ± 0.026 &  0.885 ± 0.003 \\
    \bottomrule
    \end{tabular}

    \caption{\textbf{50th percentile} evaluations for baselines on every task. Results are averaged over 8 trials, and the $\pm$ indicates the standard deviation of the reported performance. This table corresponds to the normalized performance, using the normalization methodology described in Appendix~\ref{sec:score_normalization}}.
    \label{tab:perf50b}
\end{scriptsize}
\end{table*}

\vspace{-0.1in}
\subsection{Unnormalized Experimental Results}
\label{sec:unnorm_scores}
\vspace{-0.1in}
In this section, we present the raw 100th percentile performance of the runs presented in main paper in Table~\ref{tab:perf100}. These values, presented in Table~\ref{tab:unnorm_tab}, represent the mean raw objective values and the standard deviation of the objective values attained by various offline MBO methods.

\begin{table*}[h]
\begin{scriptsize}
    \centering
\begin{tabular}{l||r|r|r|r|r|r|r}
\toprule
{} &      TF Bind 8 &     TF Bind 10 &         ChEMBL &       NAS &   Superconductor &                Ant Morphology &              D'Kitty Morphology   \\
\midrule
Auto. CbAS                    &  0.910 ± 0.044 &  0.655 ± 0.178 &      42467.285 ± 0.000 &  64.530 ± 0.764 &   77.910 ± 8.361 &   474.888 ± 44.424 &     226.156 ± 7.043 \\
CbAS                          &  0.927 ± 0.051 &  0.738 ± 0.239 &   46681.988 ± 4987.456 &  66.360 ± 0.820 &  93.078 ± 12.695 &   469.499 ± 30.570 &     209.412 ± 9.593 \\
BO-qEI                        &  0.798 ± 0.083 &  0.742 ± 0.150 &   30069.684 ± 3187.300 &  70.447 ± 0.606 &   74.322 ± 6.347 &    413.084 ± 0.000 &     213.816 ± 0.000 \\
CMA-ES                        &  0.953 ± 0.022 &  0.811 ± 0.090 &   31607.031 ± 1578.222 &  69.475 ± 0.821 &   86.072 ± 4.508 &  799.394 ± 715.702 &       4.290 ± 1.505 \\
Gradient Ascent               &  0.977 ± 0.025 &  0.762 ± 0.155 &   32514.541 ± 2612.903 &  63.770 ± 0.000 &   95.789 ± 4.436 &  -100.265 ± 22.118 &    187.206 ± 27.274 \\
Grad. Min                     &  0.984 ± 0.012 &  0.729 ± 0.126 &    32617.006 ± 370.390 &  63.770 ± 0.000 &   93.590 ± 1.719 &    80.853 ± 62.308 &    205.639 ± 13.427 \\
Grad. Mean                    &  0.986 ± 0.012 &  0.714 ± 0.071 &   33715.059 ± 1136.034 &  63.770 ± 0.000 &   92.265 ± 3.206 &    48.064 ± 78.555 &    209.355 ± 13.928 \\
MINs                          &  0.905 ± 0.052 &  0.599 ± 0.082 &   42732.578 ± 5126.862 &  66.709 ± 0.471 &   86.702 ± 4.171 &   505.515 ± 34.934 &    273.479 ± 14.184 \\
REINFORCE                     &  0.948 ± 0.028 &  0.786 ± 0.137 &   41448.012 ± 3220.380 &  39.720 ± 0.000 &   88.996 ± 2.389 &  -127.440 ± 30.831 &  -194.540 ± 238.857 \\
COMs                          &  0.945 ± 0.033 &  0.649 ± 0.153 &  391827.500 ± 2273.631 &  64.041 ± 1.431  &   81.238 ± 6.170 &   535.125 ± 16.064 &    278.344 ± 17.727 \\
\bottomrule
\end{tabular}
\caption{\textbf{Unnormalized 100th percentile} unnormalized evaluations for baselines on every task. Results are averaged over 8 trials, and the $\pm$ indicates the standard deviation of the reported performance. This table corresponds to the unnormalized performance.}
\label{tab:unnorm_tab}
\end{scriptsize}
\end{table*}

\subsection{Computation Resources}
\label{sec:computational}
The amount of computation resources required to produce the experiments in this paper is relatively modest except for the NAS tasks. We ran our experiments on a single server with 2 Intel Xeon E5-2698 v4 CPUs and 8 Nvidia Tesla V100 GPUs. All our experiments can be completed within 96 hours on this single machine.

\section{Additional MBO Tasks That Were Discarded From Our Benchmark}
\label{sec:unused_tasks}

The main benchmark consists of eight offline MBO tasks, four of which have discrete design-spaces, and four of which have contiguous design-spaces. In addition to the provided tasks, we also experimented with two other candidate MBO tasks from prior work \cite{angermuller2020p3b0, brookes2019conditioning}, but chose to not include them in the final benchmark due to lack in-distinguishable results across all methods, suggesting that these tasks may not be suitable for devising better algorithms. 

\subsection{GFP}
GFP uses the oracle function derived from \citet{RaoBTDCCAS19}. This oracle is a transformer regression model with 4 attention blocks and a hidden size of 64. The Transformer is fit to the entire hidden GFP dataset, making it possible to sample a protein design that achieves a higher score than any other protein visible to an offline MBO algorithm. Our Transformer has a Spearman's rank correlation coefficient of 0.8497 with a held-out validation set derived from the GFP dataset.

The GFP task provided is a derivative of the GFP dataset \cite{sarkisyan2016GFP}. The dataset we use in practice is that provided by \cite{brookes2019conditioning} at the url \url{https://github.com/dhbrookes/CbAS/tree/master/data}. We process the dataset such that a single training example consists of a protein represented as a tensor $\bx_{\text{GFP}}\in\{0,1\}^{237\times20}$. This tensor is a sequence of 237 one-hot vectors corresponding to which amino acid is present in that location in the protein. We use the dataset format of \cite{brookes2019conditioning} with no additional processing. The data was originally collected by performing laboratory experiments constructing proteins similar to the Aequorea victoria green fluorescent protein and measuring fluorescence. We employ the full dataset of 56086 proteins when learning approximate oracles for evaluating offline MBO methods, but restrict the training set given to offline MBO algorithms to 5000 samples drawn from between the 50th percentile and 60th percentile of proteins in the GFP dataset, sorted by fluorescence values. This subsampling procedure is consistent with prior work \cite{brookes2019conditioning}.

\subsection{UTR}

UTR uses a Transformer as the oracle function, which differs from the CNN that was originally used by \cite{angermuller2020p3b0}. Our reasoning for making this change is that the Transformer is a newer and possibly higher capacity model that may be less prone to mistakes than the shallower CNN model proposed by \citet{sample2019human}. This Transformer has 4 attention blocks and a hidden size of 64. The Transformer is fit to the entire hidden UTR dataset, making it possible to sample a DNA sequence that achieves a higher score than any other sequence visible to an offline MBO algorithm. The resulting model has a spearman's rank correlation of 0.6424 with a held-out validation set.

The UTR task is derived from work by \citet{sample2019human} who trained a CNN model to predict the expressive level of a particular gene from a corresponding 5'UTR sequence. Our use of the UTR task for model-based optimization follows \citet{angermuller2020p3b0}, where the goal is to design a length 50 DNA sequence to maximize expression level. We follow the methodology set by \citet{sample2019human} to sort all length 50 DNA sequences in the unprocessed UTR dataset by total reads, and then select the top 280,000 DNA sequences with the most total reads. The result is a dataset containing 280,000 samples of length 50 DNA sequences $\bx_{\text{UTR}}\in\{0,1\}^{50\times4}$ and corresponding ribosome loads. When training offline MBO algorithms, we subsequently eliminate the top 50\% of sequences ranked by their ribosome load, resulting in a visible dataset with only 140,000 samples.

\subsection{Additional Experimental Results}

We report the normalized performance of all baselines on the three additional MBO tasks that were not chosen for inclusion in the benchmark. Note that for GFP \cite{brookes2019conditioning} and UTR \cite{angermuller2020p3b0, sample2019human} performance of offline MBO method is not not distinguishable, and we consider this an indication each task is not suitable for benchmarking offline MBO methods. We encourage future revisions of these tasks. 

\begin{table*}[h]
\centering
\begin{tabular}{l||r|r|r}
\toprule
{} &            GFP &            UTR \\
\midrule
Auto. CbAS                    &  0.865 ± 0.000 &  0.691 ± 0.012  \\
CbAS                          &  0.865 ± 0.000 &  0.694 ± 0.010 \\
BO-qEI                        &  0.254 ± 0.352 &  0.684 ± 0.000 \\
CMA-ES                        &  0.054 ± 0.002 &  0.707 ± 0.014  \\
Grad.                         &  0.864 ± 0.001 &  0.695 ± 0.013  \\
Grad. Min                     &  0.864 ± 0.000 &  0.696 ± 0.009  \\
Grad. Mean                    &  0.864 ± 0.000 &  0.693 ± 0.010  \\
MINs                          &  0.865 ± 0.001 &  0.697 ± 0.010  \\
REINFORCE                     &  0.865 ± 0.000 &  0.688 ± 0.010  \\
COMs                          &  0.864 ± 0.000 &  0.699 ± 0.011  \\
\bottomrule
\end{tabular}
\caption{\textbf{Normalized 100th percentile} normalized evaluations for baselines on unused tasks. Each entry reports the empirical mean and empirical standard deviation over 8 independent trials.}
\label{tab:unused_task_tab}
\vspace{-0.4cm}
\end{table*}

\section{Normalization Of Inputs and Outputs Is Important for Gradient Ascent}
\label{sec:grad_normalization}

\begin{figure}[h]
\vspace{-10pt}
\centering
\includegraphics[width=0.7\columnwidth]{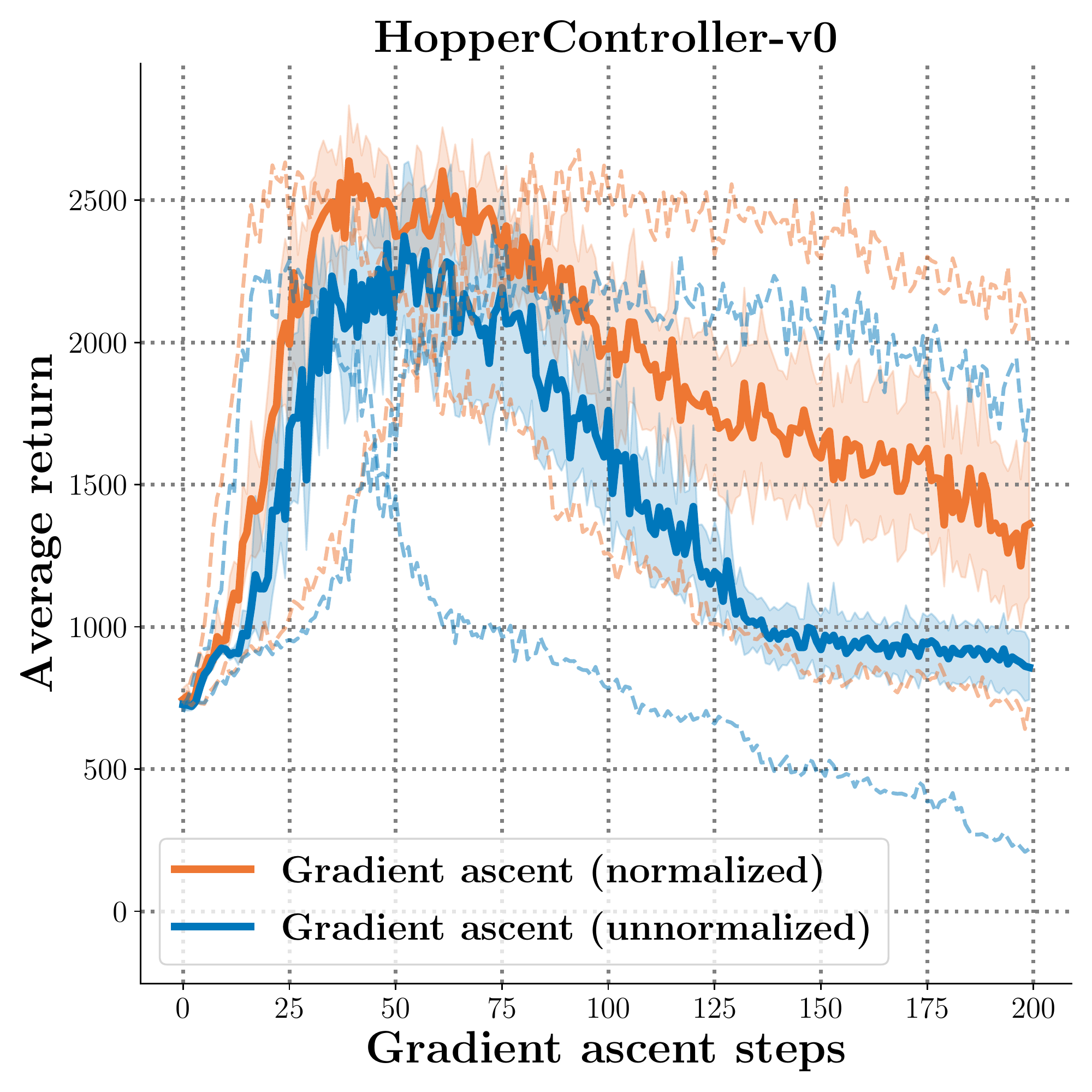}
\label{fig:plot_two_sweeps}
\vspace{-10pt}
\end{figure}

An important component for the good performance of the gradient ascent baseline is the normalization of design space. 
We found that the identical gradient-ascent baseline performed a factor 1.4x worse on Hopper Controller, when optimizing in the space of unnormalized designs and objective values, as seen in Figure~\ref{fig:plot_two_sweeps}. This indicates that normalization is key in obtaining good performance with a na\"ive gradient ascent baseline. For continuous design-space tasks, we normalize both the designs, and the scores to have unit Gaussian statistics. For discrete design-space tasks, we first map designs to real-valued logits of a categorical distribution before performing this normalization. See the \href{https://github.com/rail-berkeley/design-bench/blob/new-api/design_bench/datasets/discrete_dataset.py}{official code} for how this mapping is performed. This is a necessary part of the optimization workflow because scores vary by several orders of magnitude in the dataset, for example, 0.91 for TF Bind 8 and as high as 799.394 for Ant Morphology. The specific normalization equation is given below.
\begin{gather}
    \tilde{\bx}_{i,j} = \frac{ \bx_{i,j} - \mu(\bx_{,j}) }{ \sigma(\bx_{,j}) } \;\; : \;\; \bx \in \mathbb{R}^{N \times D}
\end{gather}
We also normalize the objective values in a similar fashion to have unit Gaussian statistics. The result in a new set of designs $\tilde{\bx}$ and objective values $\tilde{y}$ that is optimized over
\begin{gather}
    \tilde{y}_{i,j} = \frac{ y_{i,j} - \mu(y_{,j}) }{ \sigma(y_{,j}) } \;\; : \;\; y \in \mathbb{R}^{N \times 1}
\end{gather}
The gradient ascent procedure is performed in the space of these normalized designs. Suppose $T$ steps of gradient ascent have been taken, and a final normalized solution $\tilde{\bx}^{*}_{T}$ is found. This solution is de-normalized using the following transformation.
\begin{gather}
    \left( \bx^{*}_{T} \right)_{ij} = \left( \tilde{\bx}^{*}_{T} \right)_{ij} \cdot \sigma(\bx_{,j}) + \mu(\bx_{,j}) 
\end{gather}
This normalization strategy is heavily inspired by data whitening, which is known to reduce the variance of machine learning algorithms that learn discriminative mappings on that data. The learned model of the objective function is one such discriminative model, and normalization likely improves the consistency of Gradient Ascent across independent experimental trials.

\section{Hyperparameter Selection Workflow}\label{sec:hparam_select}

Hyperparameter tuning under a restricted computational budget is emerging as an import research domain in optimization \cite{Sivaprasad2020, DodgeGCSS19, Jordan2020}. Care must be taken when tuning each of the prescribed algorithms so that only offline information about the task is used for hyperparameter selection.
Formally, this means that the hyperparameters, $\mathcal{H}$, are conditionally independent of the particular value of the performance metric $\mathcal{M}$, given the offline task dataset $\mathcal{D}$. Examples of hyperparameter selection strategies that violate this requirement might, for example, perform a grid search over $\mathcal{H}$ and take the set that maximizes the performance metric, but this is not offline. An example of a tuning strategy that is fully offline is tuning the parameters of a learned model such that is is a good fit for the task dataset $\mathcal{D}$. One can choose $\mathcal{H}$ that minimizes a validation loss, such as negative log likelihood. A detailed record of hyperparameters can be found in the experiment scripts located alongside our reference implementations: \href{https://github.com/brandontrabucco/design-baselines}.

We now present guidelines for hyperparameter selection (i.e. \textbf{\textit{workflow}}) for methods evaluated in the benchmark. These are general principles that can be used to tune the hyperparameters of these methods on a new task in an offline fashion. While we only present workflow details for methods we benchmark in Section~\ref{sec:exps}, we expect that these general strategies will allow users to devise analogous schemes for tuning hyperparameters of new offline MBO methods with shared components.

\subsection{Strategy For Autofocused CbAS}\label{sec:tuning_autofocus}
The main tunable components of Autofocused methods \citep{fannjiang2020autofocused} are the learned objective function, and the generative model fit to the data distribution. When training the learned objective function, tracking a validation performance metric like rank correlation is helpful to ensure that the resulting learned model is able to generalize beyond its training dataset. This tracking is especially important for Autofocused methods because re-fitting the learned objective model during importance can lead to divergence if the importance weights generated by Autofocusing are very large or very small in magnitude. The algorithm is tuned well if, for example, the validation rank correlation stays above a positive threshold, such as a threshold of 0.9.

The second component of Autofocused methods is the fit of the generative model used for sampling designs. The algorithm has the best chance of success if the generative model can generalize beyond the dataset in which it was trained. This can be monitored by holding out a validation set and tracking a metric such as negative log likelihood on this held-out set. In the case when the generative model is not an exact likelihood-based generative model---for example, a VAE---other validation metrics can be used that measure the fit of the generative model on a validation set. The generative model is especially impacted by the importance sampling procedure used by Estimation of Distribution Algorithms (EDAs), and tracking the effective sample size of the importance weights can help diagnose when the generative model is failing to generalize to a validation set.

\subsection{Strategy For CbAS}
The main tunable components of CbAS methods \citep{brookes2019conditioning} are the learned objective function, and the generative model fit to the data distribution. While the learned objective function is not affected by the importance sampling weights generated by CbAS, the same tuning strategy described in section~\ref{sec:tuning_autofocus} that focuses on generalization to a validation set is effective. Generative model tuning can also follow an identical strategy to that described in section~\ref{sec:tuning_autofocus}, which focuses on the ability for the generative model to represent samples outside of its training set. In the case of a $\beta\text{-VAE}$, which is used with CbAS in this work, the main parameter for controlling this generalization ability is the $\beta$ parameter. We found that $\beta$ is task specific, and must be found in order for the CbAS optimizer using $\beta\text{-VAE}$ to generate samples that are in the same distribution as its validation set. This value can be tuned in practice using a validation metric like that in section~\ref{sec:tuning_autofocus}.

\subsection{Strategy For MINs}\label{sec:tuning_mins}

The main tunable components of MINs~\citep{kumar2019model} are the learned objective function, and the generative model fit to the data distribution. The learned objective function is typically trained using a maximum likelihood objective, and the validation log-likelihood (or regression error) can be directly tracked. The learned objective function should train until  a minimum validation loss is reached, which ensured that the model will generalize as well as possible beyond its training set. Since only the static task dataset is used for this---it may be split into train/validation sets---this tuning strategy is fully offline.

The generative model for MINs is an inverse mapping $\mathbf{x} = f^{-1}(y, \mathbf{z})$, conditioned on the objective value $y$. Training conditional generative models is considerable less stable than unconditional generative models, so in addition to monitoring the fit of a validation set recommended in section~\ref{sec:tuning_autofocus}, it is also necessary to track the extent of the dependence of the generative model's predictions on the objective value $y$. This can be evaluated in practice by comparing the distribution of $x$ from the conditional generative model $p(\bx | y)$ to an unconditional generative model $p(\bx)$ with an identical initialization, or by comparing if $p(\bx|y)$ is independent of $y$ by querying the inverse model for different values of $y$ and visualizing the similarity in the predictions of $\bx$. One metric for more formally studying the extent of the dependence of $\bx$ on $\mathbf{z}$ is the mutual information $I(\bx; \mathbf{z})$. The conditional generative model has an appropriate fit if for some positive threshold $c$ we have that $I(\bx; \mathbf{z}) > c$.

\subsection{Strategy For Gradient Ascent}\label{sec:tuning_gradient_ascent}
The main tunable components of Gradient Ascent MBO methods are the learned objective function, and the parameters for gradient ascent. The learned objective function is typically trained using a maximum likelihood objective under a Gaussian distribution, and the methodology for obtaining a high-performing learned objective function is identical to that in section~\ref{sec:tuning_mins}. The second aspect of gradient ascent MBO algorithms are the parameters of the gradient-based optimizer for the designs---such as its learning rate, and the number of gradient steps it performs. The learning rate should be small enough that the gradient steps taken increase the prediction of the learned objective function---if the learning rate is too large, gradient steps may not follow the path of steepest ascent of the objective function. The number of gradient steps is more difficult to tune. The strategy we used is a fixed number of steps, and an offline criterion to select this parameter is future work.

\subsection{Strategy For REINFORCE}

The main tunable components of REINFORCE-based MBO methods are the learned objective function, and the parameters for the policy gradient estimator. The learned objective function is typically trained using a maximum likelihood objective, and the methodology for obtaining a high-performing learned objective function is identical to that in section~\ref{sec:tuning_mins}. The remaining parameters to tune are specific to REINFORCE. The distribution of the policy should be carefully selected to be able to model the distribution of designs. For continuous MBO tasks, a Gaussian distribution is appropriate, and for discrete MBO tasks, a categorical distribution is appropriate. In addition, the learning rate, and optimizer should be selected so that policy updates improve the model-predicted score.

\subsection{Strategy For Bayesian Optimization}

The main tunable components of Bayesian Optimization MBO methods \citep{botorch2019} are the learned objective function, and the parameters for the bayesian optimization loop. The learned objective function is typically trained using a maximum likelihood objective, and the methodology for obtaining a high-performing learned objective function is identical to that in section~\ref{sec:tuning_mins}. For a detailed review of the strengths and weaknesses of various Bayesian Optimization strategies and their hyperparameters, we refer the reader to the BoTorch documentation, available at the BoTorch website \url{https://botorch.org/docs/overview}. In this work we employ a Gaussian Process as the model, and the quasi-Monte Carlo Expected Improvement acquisition function, which has the advantage of scaling up to our high-dimensional optimization problems. 

\subsection{Strategy For Covariance Matrix Adaptation (CMA-ES)}

The main tunable components of Covariance Matrix Adaptation MBO methods are the learned objective function, and the parameters for the evolution strategy. The learned objective function is typically trained using a maximum likelihood objective, and the methodology for obtaining a high-performing learned objective function is identical to that in Subsection~\ref{sec:tuning_mins}. For a detailed review of the strengths and weaknesses of various Bayesian Optimization strategies and their hyperparameters, we refer the reader to an open-source implementation of CMA-ES and its corresponding documentation \url{https://github.com/CMA-ES/pycma}. In this work we employ the default settings for CMA-ES reported in this open source implementation, with $\sigma = 0.5$.

\subsection{Strategy For Conservative Objective Models (COMs)}

Conservative Objective Models has three main tunable parameters, and we refer the reader to the original paper for a full experimental description \cite{trabuccoconservative}. The first parameter for COMs is the degree to which the objective model is allowed to overestimate the objective value for off-manifold designs. This parameter can be implemented as a constraint with threshold $\tau$, or as a penalty with weight $\alpha$. This parameter is chosen to be as high as possible, permitting high validation performance. When either $\tau$ or $\alpha$ imposes too much conservatism, this regularizes the objective model, and may lead the model to poorly fit the dataset $\dataset$. This parameter is uniformly chosen to be 2 for all discrete tasks and 0.5 for all continuous tasks. The second tunable parameter of COMs is the number of gradient ascent steps to perform when optimizing $\bx$, and is uniformly chosen to be 50. The final parameter is the learning rate used when optimizing $\bx$, which is uniformly chosen to be $2\sqrt{d}$ for all discrete tasks and $0.05 \sqrt{d}$ for all continuous tasks, where $d$ is the cardinality of the design space.

\end{appendices}

\end{document}